\documentclass[10pt]{article} 

\usepackage{multirow}

\usepackage[accepted]{tmlr}
\usepackage{graphicx}
\usepackage{xcolor}
\usepackage{caption}
\usepackage{hyperref}
\usepackage{booktabs}
\usepackage{colortbl}
\usepackage{amsmath}
\usepackage{amssymb}
\usepackage{amsfonts} 
\usepackage{multirow}
\usepackage{color}
\usepackage[dvipsnames]{xcolor}
\usepackage{makecell}
\usepackage{pifont}
\usepackage{float}
\usepackage{etoc}

\newcommand{\cmark}{\ding{51}} 
\newcommand{\xmark}{\ding{55}} 
\usepackage{tabularx}

\usepackage{amsmath,amsfonts,bm}

\def\eqref#1{equation~\ref{#1}}

\def\1{\bm{1}}

\DeclareMathAlphabet{\mathsfit}{\encodingdefault}{\sfdefault}{m}{sl}
\SetMathAlphabet{\mathsfit}{bold}{\encodingdefault}{\sfdefault}{bx}{n}

\usepackage{hyperref}
\usepackage{url}
\usepackage{xcolor}
\definecolor{MidPurple}{HTML}{855E95} 
\definecolor{darkblue}{RGB}{0,0,120}

\hypersetup{
    colorlinks=true,
    citecolor=darkblue,
    linkcolor=darkblue,
    urlcolor=darkblue
}
\title{Cog-VADU: A Training-Free Cognitive Reasoning\\
Framework for Video Anomaly Detection and Understanding}

\author{
\name Mohd Ubaid Wani \email m.wani@surrey.ac.uk \\
\addr Centre for Vision, Speech and Signal Processing (CVSSP)\\
University of Surrey, UK
\AND
\name Sara Atito\textsuperscript{*} \email sara.atito@surrey.ac.uk \\
\addr Surrey Institute for People-Centred AI (PAI)\\
University of Surrey, UK
\AND
\name Josef Kittler \email j.kittler@surrey.ac.uk \\
\addr Centre for Vision, Speech and Signal Processing (CVSSP)\\
University of Surrey, UK
\AND
\name Muhammad Awais \email muhammad.awais@surrey.ac.uk \\
\addr Surrey Institute for People-Centred AI (PAI)\\
University of Surrey, UK
}

\def\month{08}
\def\year{2026}
\def\openreview{\url{https://openreview.net/forum?id=QcuSMNG7J8}}

\begin{document}

\maketitle
\noindent
$^{*}$Corresponding author: Sara Atito
(\href{mailto:sara.atito@surrey.ac.uk}{sara.atito@surrey.ac.uk}).
\begin{abstract}
Video Anomaly Detection (VAD) aims to temporally localize abnormal events in videos. 
Most existing approaches rely on dataset-specific training and curated annotations, limiting generalization in open-set scenarios. 
Recent zero-shot methods based on Large Vision-Language Models (LVLMs) alleviate this dependency but often lack temporal continuity and structured reasoning. We propose \textbf{Cog-VADU}, a fully training-free framework that reformulates VAD as a sequential cognitive reasoning task. 
Cog-VADU introduces \emph{Chain-of-Anomaly Detection Thought Prompting} (CoADTP), which unrolls an LVLM into a recurrent reasoning chain across video segments. 
By propagating structured rationales over time, the model maintains implicit temporal memory, enabling robust discrimination between complex anomalies and high-motion normal activities. To improve reliability, we further design a cross-modal re-ranking stage that aligns textual rationales with visual embeddings, enforcing semantic consistency and temporal coherence for refined and stable predictions. Extensive experiments on multiple public VAD benchmarks demonstrate that Cog-VADU achieves competitive zero-shot performance. Moreover, cross-model evaluations show that CoADTP consistently enhances reasoning-based anomaly detection in a model-agnostic manner, providing interpretable and generalizable anomaly understanding for real-world applications.
\end{abstract}

\section{Introduction}
\label{sec:intro}

Video surveillance plays a vital role in monitoring activities in public and private spaces.
\emph{Video Anomaly Detection (VAD)}~\cite{sultani2019realworldanomalydetectionsurveillance} seeks to temporally localize unusual events such as violence, accidents, or security violations. 
Anomalies are typically characterized as deviations in appearance or motion from expected patterns~\cite{5562666}, often confined to short segments. 
However, abnormality is not purely statistical. 
Events may appear anomalous due to contextual mismatch rather than visual irregularity alone. 
A running crowd, for instance, may indicate panic in one setting and celebration in another. 
Such distinctions require situational understanding that extends beyond low-level motion cues.


\noindent\textit{\textbf{Early vision-only methods}}~\cite{liu2018futureframepredictionanomaly,10.1145/3343031.3350998,wu2020looklistenlearningmultimodal,tian2021weaklysupervisedvideoanomalydetection,wan2021weaklysupervisedvideoanomaly,9591368,zhou2023dualmemoryunitsuncertainty,lv2023unbiasedmultipleinstancelearning,zhang2024glancevadexploringglancesupervision} primarily focus on modeling appearance and motion patterns to compute anomaly scores from video data. 
While these approaches have shown effectiveness in constrained or well-defined environments, they fundamentally lack semantic and contextual reasoning capabilities. 
Consequently, they are unable to explain \emph{who} is involved, \emph{what} is occurring, or \emph{why} an event should be considered anomalous. 
The absence of high-level contextual understanding, an essential component for reliable video anomaly detection, significantly limits their robustness and generalization when deployed in complex real-world surveillance scenarios.
\begin{figure}[htbp]
\centering
\captionsetup{font=scriptsize, skip=2pt}
\includegraphics[width=1.0\textwidth]{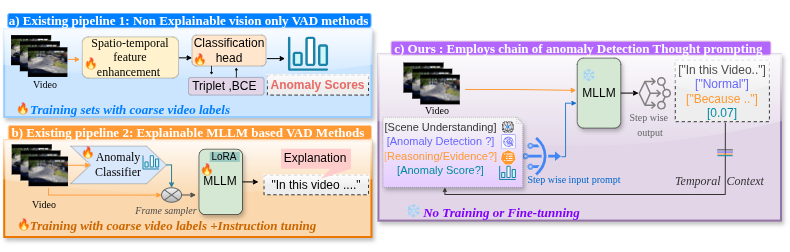}

\caption{
Comparison of video anomaly detection paradigms. 
(a) Vision-only methods rely on supervised training and produce anomaly scores without semantic reasoning. 
(b) Multimodal LLM-based VAD generates anomaly confidence and provides post-hoc explanations. 
(c) Cog-VADU employs a frozen LVLM with structured prompting and temporal cues to perform joint, training-free anomaly detection and reasoning.
}
\label{fig:overview}
\end{figure}

\noindent\textit{\textbf{Multimodal approaches}}~\cite{yuan2023surveillancevideoandlanguageunderstandingnew,wu2023vadclipadaptingvisionlanguagemodels,10208872,yang2024textpromptnormalityguidance,tang2024hawklearningunderstandopenworld,ma2025sherlockmultiscenevideoabnormal,yang2025monitorexploitinglargelanguage} seek to overcome these limitations by incorporating multimodal cues through vision-language models (VLMs). 
By integrating visual observations with semantic and linguistic information, these methods improve detection performance and enhance interpretability, particularly in surveillance scenarios where visual cues alone are ambiguous or insufficient. 
Despite their promising results, the majority of these approaches rely on task-specific fine-tuning or instruction tuning using large-scale, carefully curated datasets, which are expensive to collect and difficult to scale. 
Such dependence on extensive training not only increases computational cost but also necessitates retraining when deployed in new environments or scenarios with shifting contextual distributions, thereby limiting scalability and adaptability,  constraining open-set generalization.

To reduce the reliance on training, several training-free approaches, such as~\cite{zanella2024harnessinglargelanguagemodels}, exploit pretrained VLMs for caption generation and semantic matching, often employing models such as BLIP-2~\cite{li2023blip2bootstrappinglanguageimagepretraining}. 
Although these methods eliminate the need for additional training, they typically rely on shallow captioning or similarity-based matching strategies. 
As a result, they lack explicit multi-step reasoning mechanisms and struggle to capture complex contextual relationships, long-range temporal dependencies, and evolving scene dynamics. 
This often causes them to miss nuanced temporal and situational cues that are critical for distinguishing subtle anomalies from normal activities, particularly in cases requiring semantic interpretation and temporal causality.

In parallel, other approaches based on large vision-language models (LVLMs)~\cite{tang2024hawklearningunderstandopenworld,huang2025vadr1videoanomalyreasoning,zhu2025vaur1advancingvideoanomaly,ye2025veraexplainablevideoanomaly} attempt to further exploit the inherent reasoning capabilities of such models to provide more interpretable anomaly detection, frequently by introducing explicit reasoning chains or language-based explanations. 
However, despite incorporating reasoning components, these methods still depend on task-specific training or supervision and, in some cases, specialized reasoning datasets with the requirement of high computational resources.
This reliance restricts their ability to generalize across diverse datasets, unseen anomaly types, and real-world surveillance scenarios where annotations are limited or entirely unavailable.

\noindent \textbf{Key Findings.}
\noindent Fundamentally, video anomaly detection requires \emph{contextual reasoning} rather than relying solely on pattern deviation or statistical irregularities. 
Human judgment of anomalies involves understanding the overall scene, interpreting situational context, modeling temporal evolution, and reasoning about deviations from expected behavior before arriving at a final decision. 
Such reasoning enables the distinction between visually similar yet semantically different events, such as staged wrestling versus genuine acts of violence, or crowded yet orderly gatherings versus chaotic confrontations. 
Despite its critical importance, this reasoning-centric perspective remains largely underexplored in existing VAD methods, which predominantly depend on low-level visual deviations, heuristic thresholds, or naive prompting strategies that lack structured reasoning and temporal awareness.
In light of these observations, we investigate the following unexplored yet fundamental question:
\begin{quote}
\emph{Can Video Anomaly Detection (VAD) be reformulated as a simple reasoning-driven, zero-shot detection and understanding problem by leveraging pretrained vision-language models, without any task-specific training or fine-tuning?}
\end{quote}

We address this question with \textbf{Cog-VADU}, illustrated in Fig.~\ref{fig:overview}. 
Cog-VADU treats anomaly detection as a structured cognitive reasoning process unfolding across video segments. 
A frozen, model LVLM is prompted to generate step-wise rationales that persist over time, forming an implicit temporal memory. 
Initial predictions are then refined through cross-modal rationale alignment, enforcing semantic consistency and suppressing hallucinations. 
The resulting system produces temporally coherent anomaly scores along with interpretable, glass-box explanations without any fine-tuning.

\paragraph{Contributions.}
\begin{itemize}
\renewcommand{\labelitemi}{$\bullet$}
\item We present \textbf{Cog-VADU}, a training-free formulation of VAD that models anomaly detection as sequential cognitive reasoning and understanding over video segments using pretrained LVLMs.
\item We propose \emph{Chain-of-Anomaly Detection Thought Prompting (CoADTP)}, enabling step-wise rationale propagation and temporally coherent zero-shot inference.
\item We introduce a cross-modal re-ranking rationale refinement stage that aligns textual reasoning with visual representations to stabilize predictions.
\item We provide unified re-annotations of UCF-Crime and XD-Violence that account for causal and temporal relationships, reducing label inconsistencies and enabling fair evaluation.
\item Experiments across benchmarks demonstrate competitive zero-shot performance alongside interpretable reasoning outputs.
\end{itemize}

\section{Related Work}
\label{sec:related_work}
\textbf{Video Anomaly Detection (VAD).}
Video Anomaly Detection (VAD) aims to identify frames or temporal segments that deviate from normal behavior~\cite{hasan2016learningtemporalregularityvideo}, and is fundamental to public safety~\cite{sultani2019realworldanomalydetectionsurveillance} and video understanding~\cite{wu2020looklistenlearningmultimodal}. 
Traditional VAD methods predominantly rely on deep neural networks using visual cues, achieving strong performance but lacking semantic understanding and interpretability.
Existing methods are commonly categorized as Non-Explainable, such as unsupervised, weakly supervised, and fully supervised 
and Explainable, like MLLM-VADs.

\noindent\textit{\textbf{Unsupervised Methods.}}
Unsupervised VAD~\cite{liu2018futureframepredictionanomaly,gong2019memorizingnormalitydetectanomaly,wang2019godsgeneralizedoneclassdiscriminative,thakare2022dyannetscenedynamicityguided,tur2023exploringdiffusionmodelsunsupervised,yang2023videoeventrestorationbased} learn normal patterns from unlabeled or normal-only data and detects anomalies as deviations. 
While effective for modeling regular behavior, they struggle with subtle anomalies and generalization to unseen environments due to the absence of anomaly priors.

\noindent\textit{\textbf{Fully Supervised Methods.}}
Fully supervised approaches~\cite{wang2010crowd,10.1145/3343031.3350998} utilize frame-level annotations for both normal and abnormal events, enabling precise localization but requiring extensive manual labeling, which limits scalability.

\noindent\textit{\textbf{Weakly Supervised Methods.}}
Weakly supervised VAD~\cite{sultani2019realworldanomalydetectionsurveillance,tian2021weaklysupervisedvideoanomalydetection,lv2023unbiasedmultipleinstancelearning} leverages video-level labels, striking a balance between annotation cost and performance.
The dominant paradigm in this setting is Multiple Instance Learning (MIL), first introduced to VAD by \citet{sultani2019realworldanomalydetectionsurveillance}, where a video is treated as a bag of segments, and a positive (anomalous) bag is assumed to contain at least one anomalous segment, while a negative bag contains none.
Building on this formulation, subsequent works have progressively improved different components of the MIL framework.
These include enhanced instance weighting~\cite{ilse2018attentionbaseddeepmultipleinstance}, improved feature magnitude modeling~\cite{chen2022mgfnmagnitudecontrastiveglanceandfocusnetwork}, as well as advances in efficiency and temporal modeling~\cite{gao2025steadspatiotemporalefficientanomaly}.
More recent efforts further address robustness to label noise and uncertainty~\cite{zhou2023batchnormbasedweaklysupervisedvideo,zhou2023dualmemoryunitsuncertainty}.

\noindent\textbf{Multimodal and Explainable VAD.}
To address the lack of semantic understanding, recent methods \cite{yuan2023surveillancevideoandlanguageunderstandingnew} integrate vision-language models (VLMs) to enable explainable video anomaly detection.
Early approaches such as ~\cite{zhang2024holmesvadunbiasedexplainablevideo} and its extension~\cite{zhang2025holmesvaulongtermvideoanomaly} combine temporal sampling ~\cite{zhou2023dualmemoryunitsuncertainty} with VLM-based reasoning to generate interpretable anomaly descriptions.

Zero-shot and multimodal methods have further explored pretrained VLMs.
~\cite{zanella2024harnessinglargelanguagemodels} employs BLIP-2~\cite{li2023blip2bootstrappinglanguageimagepretraining} and LLaMA-2 for caption-based anomaly scoring using task-specific prompts.
VAD-LLaMA~\cite{lv2024videoanomalydetectionexplanation} introduces instruction tuning and temporal memory for weakly supervised detection.
CLIP-based approaches such as VadCLIP~\cite{wu2023vadclipadaptingvisionlanguagemodels} perform frozen vision–language alignment enhanced with adaptive prompts and MIL-style objectives.
VERA~\cite{ye2025veraexplainablevideoanomaly} performs coarse-to-fine anomaly reasoning via scene-temporal integration, but relies on verbal reasoning training and exhibits limited generalization.

Other multimodal extensions include {TEVAD}~\cite{10208872}, which fuses visual and textual representations with multi-scale temporal modeling for caption-enhanced anomaly reasoning. Similarly, {Pu et al.}~\cite{pu2024learningpromptenhancedcontextfeatures} propose Prompt-Enhanced Learning (PEL) and Temporal Context Aggregation (TCA) to refine semantic alignment, reuse similarity matrices efficiently, and improve discriminability via knowledge-guided prompts, but still depend on training or dataset-specific design choices.

\noindent\textbf{Chain-of-Thought and Reasoning.}
Chain-of-thought (CoT) prompting~\cite{zhang2022automaticchainthoughtprompting,wei2023chainofthoughtpromptingelicitsreasoning} has demonstrated that providing exemplars with intermediate reasoning steps elicits complex reasoning in sufficiently large language models, enhancing performance on arithmetic, common sense, and symbolic tasks. Inspired by
CoT prompting in text-based LLMs.
Recent studies extend CoT to multimodal and video understanding to focus on generating intermediate
reasoning steps for visual understanding ~\cite{zhang2024improvevisionlanguagemodel,zhang2025videocotcomprehensivedatasetspatiotemporal,fei2024videoofthoughtstepbystepvideoreasoning,arnab2025temporalchainthoughtlongvideo}, yet its application to video anomaly detection remains largely unexplored.
Most existing VAD methods do not integrate structured reasoning with temporal feedback, leaving a gap between reasoning-driven prompting and practical surveillance analysis.

\noindent\textbf{Summary.} These limitations highlight the need for a training-free, zero-shot video anomaly detection framework that can fully exploit the intrinsic world knowledge, understanding, and multimodal reasoning capabilities of large vision-language models, without relying on fine-tuning or instruction-tuned datasets.

\section{Our Methodology: Cog-VADU}
\label{sec:our_method}
We introduce \textbf{Cog-VADU}, a two-stage training-free, cognitive reasoning-driven framework for video anomaly understanding and detection, built upon a frozen large vision-language model (LVLM). 
Cog-VADU performs anomaly localization, explanation, and structured cognitive reasoning in a fully zero-shot manner.
Our approach requires no fine-tuning, adapter modules, spatio-temporal backbones, or task-specific training.
In the following, we first formalize the video anomaly detection problem and then describe the design of our reasoning-driven framework.

\subsection{Problem Formulation}
Consider a test video $V = \{I_1, I_2, \ldots, I_T\}$, where $I_t$ is the $t$-th frame and $T$ is the total number of frames. 
Our goal is to predict a frame-level anomaly score sequence 
\begin{equation}
\hat{A} = [\hat{a}_1, \ldots, \hat{a}_T], \quad \hat{a}_t \in [0,1],
\end{equation}
where higher values indicate greater abnormality. 

Unlike conventional methods, Cog-VADU does not require labeled training data. 
Using a frozen, model-agnostic LVLM, it performs zero-shot inference while maintaining temporal coherence across frames. 
For each segment, the framework outputs not only anomaly scores but also interpretable semantic descriptions and structured reasoning rationales, forming a glass-box explanation grounded in visual content and temporal context.

\subsection{How Effective Is Reasoning-Based Prompting for VAD?}
\vspace{-1mm}

Prompting is central to exploiting large vision-language models (LVLMs), as it defines task intent and guides inference.
Its effectiveness depends on how well task-specific knowledge and reasoning structures are encoded.

\noindent Prior work highlights the impact of prompt design for anomaly detection.
For instance, \citet{ye2025veraexplainablevideoanomaly} learn task-specific prompts while keeping the backbone model frozen, but requires dataset-specific optimization and additional training.
In contrast, \citet{zanella2024harnessinglargelanguagemodels} uses lightweight caption-based prompting, which is efficient but lacks explicit task decomposition and structured reasoning.

\begin{figure}[H]
  \centering
  \begin{minipage}[t]{0.5\textwidth}
    \centering
    \includegraphics[width=\linewidth]{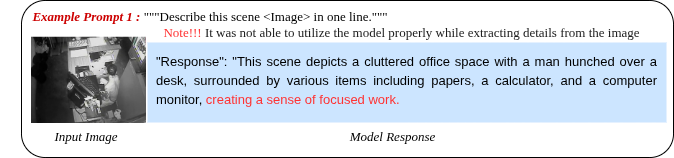}
    \caption*{(a) Prompt 1}
  \end{minipage}\hfill
  \begin{minipage}[t]{0.5\textwidth}
    \centering
    \includegraphics[width=\linewidth]{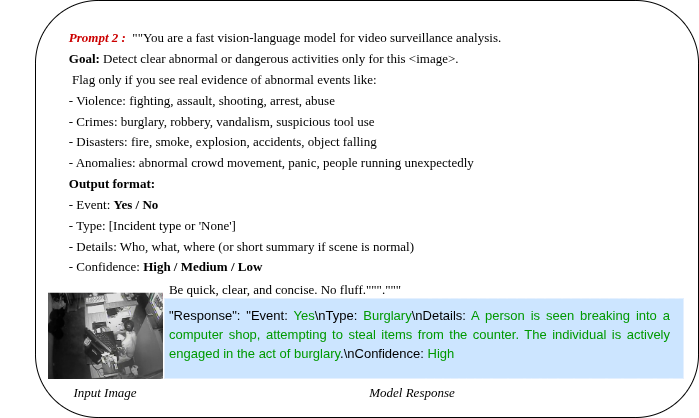}
    \caption*{(b) Prompt 2}
  \end{minipage}
\caption{Comparison of prior-aware prompts for zero-shot VAD.}
\label{fig:prompt_side_by_side_minipage}
\end{figure}

We evaluate different prompting strategies to assess their effectiveness.
A naive captioning prompt (Prompt 1) produces fluent descriptions but weak anomaly discrimination due to the lack of task guidance.
Adding structured prompts with anomaly categories and confidence scores (Prompt 2) improves performance by enforcing task awareness, as shown in Fig.~\ref{fig:prompt_side_by_side_minipage}.
However, predictions remain unstable, reflected in lower Acc@0.5 and AP, as the model overemphasizes isolated cues without contextual severity.
Moreover, such prompts lack (i) step-wise reasoning, (ii) continuous severity modeling, and (iii) temporal consistency across segments.

To address these limitations, we propose \emph{Chain-of-Anomaly Detection Thought Prompting} (CoADTP), which introduces structured reasoning, severity-aware scoring, and temporal feedback via previous segments.
As shown in Table~\ref{tab:Prompt_ablation}, CoADTP achieves the best performance, demonstrating that explicit reasoning with temporal continuity is critical for robust VAD.

\subsection{CoADTP: Chain of Anomaly-Detection Thought Prompt}
\label{subsec}

To fully exploit the reasoning capabilities of LVLMs, we reformulate video anomaly detection as a training-free, prompt-driven reasoning task.
Inspired by chain-of-thought prompting~\cite{zhang2022automaticchainthoughtprompting,wei2023chainofthoughtpromptingelicitsreasoning,fei2024videoofthoughtstepbystepvideoreasoning,arnab2025temporalchainthoughtlongvideo}, we introduce \textbf{Chain-of-Anomaly Detection Thought Prompting (CoADTP)}, which integrates anomaly priors and temporal context within a unified prompting framework.

The prompt is designed to enforce a structured reasoning pipeline prior to prediction.
Given a video clip $v_i$, the LVLM first performs scene understanding by identifying objects, actions, and interactions.
It then incorporates semantic priors from the anomaly database together with contextual information from the previous clip.
Conditioned on this combined information, the model generates structured reasoning outputs, including observations, a concise description, anomaly assessment, and a rationale explaining the decision.
Finally, it evaluates the severity of the event and produces a calibrated anomaly score.

Formally, CoADTP guides a frozen LVLM to generate outputs through three complementary components: 
(i) step-wise chain-of-thought reasoning, 
(ii) a lightweight text-based anomaly database encoding prior knowledge of normal and abnormal events, and 
(iii) temporal feedback across consecutive clips for score stability. 
These components enable consistent anomaly scoring and interpretable predictions without any additional training.

\begin{figure}[htbp]
\centering
\captionsetup{font=normalsize, skip=2pt}

\includegraphics[width=0.95\textwidth]{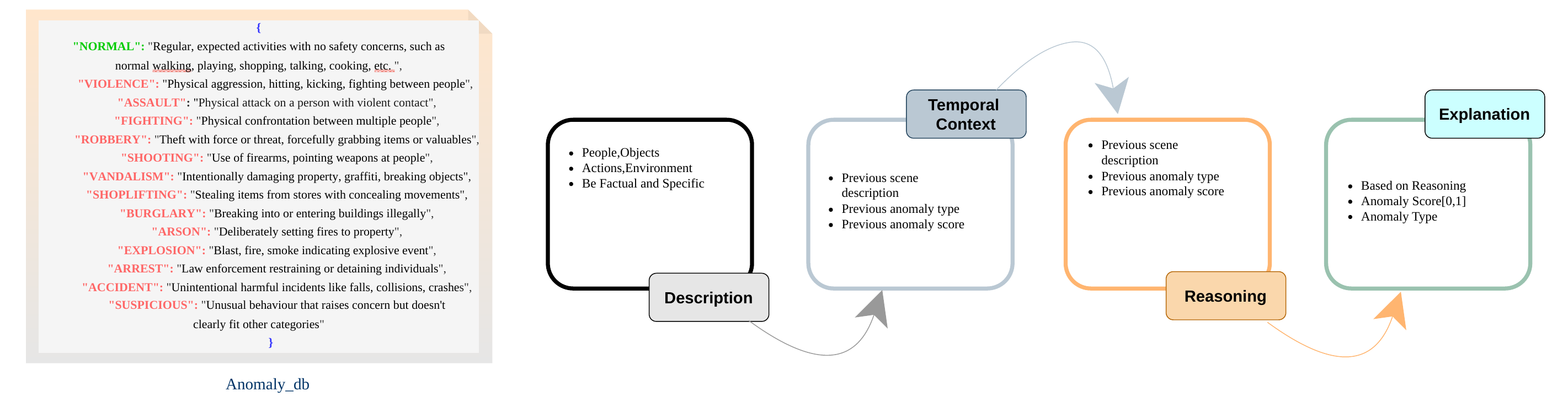}
\caption{Workflow of CoADTP integrating anomaly priors and temporal reasoning cues.}
\label{fig:my_cot_prompt}
\end{figure}

In contrast, CoADTP directly exploits the intrinsic reasoning capability of foundation-scale VLMs through structured prompting. 
The model is guided to sequentially \emph{describe}, \emph{analyze}, \emph{decide}, and \emph{justify/score} each event using prior knowledge and temporal context. 
This transforms VAD into an interpretable, context-aware inference process capable of capturing causal relations, temporal dependencies, and situational nuances entirely without training.
\paragraph{Design components.}
\begin{itemize}
\renewcommand{\labelitemi}{$\bullet$}

\item \textbf{Step-wise (chain-of-thought) reasoning.} 
Instead of directly predicting abnormality, the model follows a fixed reasoning chain:
\[
\textit{observe} \rightarrow
\textit{describe} \rightarrow
\textit{reason \& assess severity} \rightarrow
\textit{decide category} \rightarrow
\textit{score}.
\]
The final anomaly score lies in $[0,1]$, where 0 indicates normal and 1 indicates abnormal behavior. 
This decomposition breaks down complex judgments into interpretable sub-steps (Fig.~\ref{fig:my_cot_prompt}, panel~(b)).

\item \textbf{Anomaly text database (zero-shot prior).}
To inject task-specific knowledge without training, we introduce a compact \emph{verbal anomaly database} containing anomaly class names and concise definitions of normal and abnormal patterns.
By explicitly referencing this database in the prompt, the LVLM can semantically align visual evidence with anomaly categories and output both an anomaly \emph{type} and a concise rationale (Fig.~\ref{fig:my_cot_prompt}, panel~(a)). 
This eliminates the need for dataset-specific fine-tuning or annotation-heavy preparation.

\item \textbf{Temporal feedback for stability.}
To ensure temporal consistency, the prompt $P_i$ for clip $v_i$ is conditioned on the structured contextual rationale of the previous clip:
\begin{equation}
R_{i-1} =
\big(
\mathrm{desc}_{i-1},\;
\mathrm{reason}_{i-1},\;
\mathrm{type}_{i-1},\;
a_{i-1}
\big),
\end{equation}
\begin{equation}
P_i = \mathcal{P}(v_i, R_{i-1}),
\end{equation}
where $a_{i-1}$ is the anomaly score.
This recursive conditioning reduces prediction noise and improves temporal stability.
\end{itemize}

\noindent
This design leads to improved score stability and detection accuracy compared to simple prompting strategies (see Sec.~\ref{sec:experiments}).

\subsection{Raw Stage}
\label{subsec:raw}
\begin{figure}[htbp]
\centering
\captionsetup{font=normalsize, skip=5pt}
\includegraphics[width=1.0\textwidth]{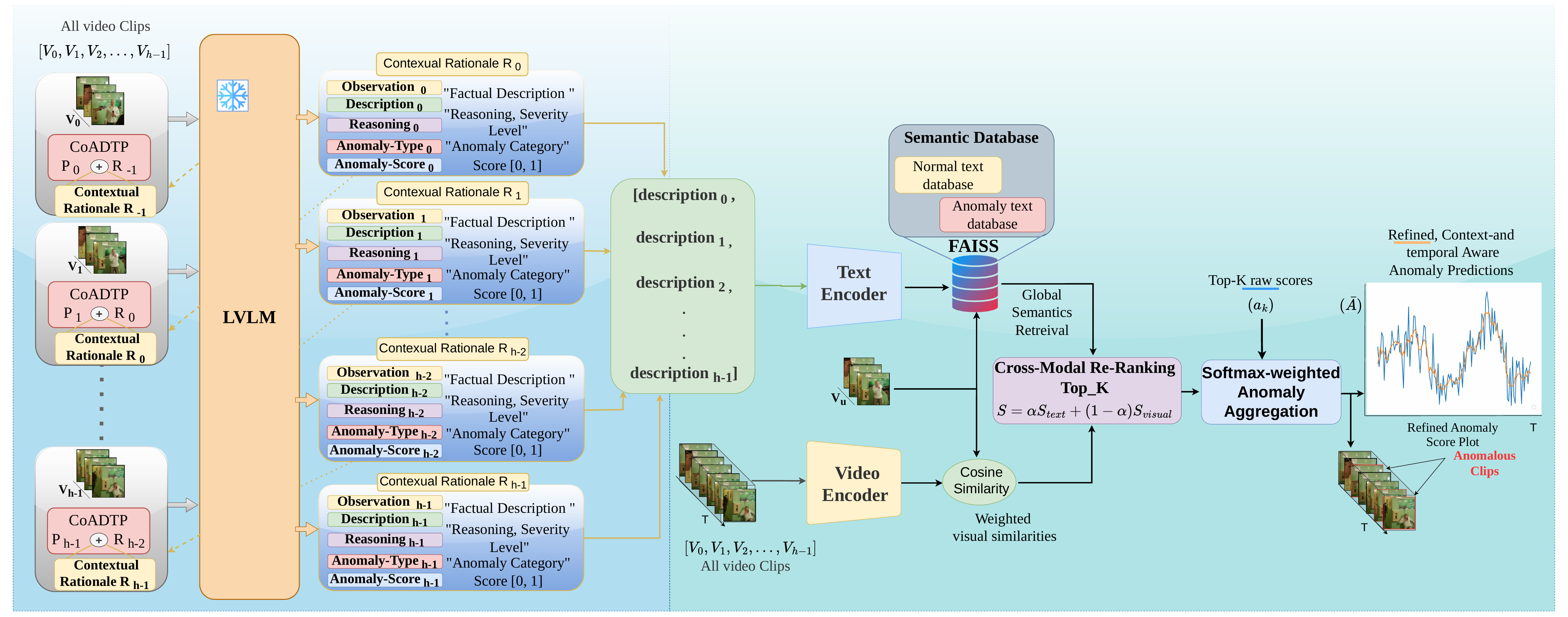}
\caption{
Overview of \textbf{Cog-VADU}, a training-free, reasoning-driven video anomaly detection framework.
Videos are divided into clips and processed by a frozen $f_{\mathrm{LVLM}}$ using
{Chain-of-Anomaly-Detection Prompting (CoADTP)} to produce anomaly scores and textual rationales.
A refinement stage employs \textbf{ImageBind} embeddings to retrieve and re-rank semantically related
visual-text contexts, yielding final contextualized anomaly scores via softmax-weighted similarity aggregation.
}

\label{fig:Cog-VADU_MODEL}
\end{figure}

Our methodology follows a two–stage design, as shown in Fig.~\ref{fig:Cog-VADU_MODEL}. 
The \emph{raw stage} produces zero-shot, segment-level anomaly scores and concise textual rationales using a vision--language model (LVLM), guided by our CoADTP prompt. No additional heads, fine-tuning, or parameter updates are used.
\textit{Segmentation and sampling.}
Let a video consist of \(F\) frames \(V=\{I_i\}_{i=1}^{F}\), where \(I_i\) denotes the \(i\)-th frame. 
We define segment centers at a fixed stride \(d\), yielding
\begin{equation}
\mathcal{C} = \{ I_{(u-1)d+1} \mid u = 1, \dots, h \}, \quad h = \left\lfloor \frac{F}{d} \right\rfloor.
\label{eq:centers}
\end{equation}
For each center frame \(I_{(u-1)d+1}\), we construct a temporal window of length \(T\) seconds (e.g., \(T=10\)s) centered around it. 
Assuming a frame rate of \(r\) frames per second, this corresponds to a window of \(Tr\) frames. 
From this window, we uniformly sample \(m=10\) frames to form the segment \(V_u\).\noindent{Prompted VLM inference.}
For each segment \(u\), we form a CoADTP prompt \(P_u\) that enforces a step-wise analysis (describe \(\rightarrow\) reason \(\rightarrow\) decide \(\rightarrow\) explain) and references a lightweight anomaly \emph{text} database; it includes temporal feedback from the previous segment (summary and score). With frozen $f_{\mathrm{LVLM}}$ parameters \(\theta\), we query
\begin{equation}
    \label{eq:initial}
\big(\text{desc}_u,\ \text{reason}_u,\ \text{type}_u,\ {a}_u\big)
= f_{\mathrm{LVLM}}\!\big(V_u;\ \theta,\ P_u\big),
\end{equation}
\indent where \(\text{desc}_u\) is a two-sentence clip description, \(\text{reason}_u\) is a brief rationale, \(\text{type}_u\) is a single anomaly category (e.g., \texttt{NORMAL}, \texttt{FIGHTING}, \texttt{ACCIDENT}), and \({a}_u\in[0,1]\) is the raw anomaly score (higher is more anomalous).
\paragraph{Outputs of the raw stage.}
Repeating \eqref{eq:initial} for all segments yields
\begin{equation}
\begin{aligned}
\tilde{\mathbf{A}}&=[a_1,\dots,a_h], &
\mathbf{T}&=[\text{type}_1,\dots,\text{type}_h], \\
\mathbf{D}&=[\text{desc}_1,\dots,\text{desc}_h], &
\mathbf{R}&=[\text{reason}_1,\dots,\text{reason}_h].
\end{aligned}
\label{eq.6}
\end{equation}
\noindent These raw scores and textual outputs are passed to the refinement stage, where we stabilise and sharpen localisation and inject global scene context using cross-model context retrieval (Sec.~\ref{subsec:Analysis}). Quantitative analyses of the raw-stage behaviour for different prompting strategies are reported in the Table ~\ref{tab:Prompt_ablation}.

\paragraph{Prompt Sensitivity Analysis.}
To verify that Cog-VADU's raw-stage performance is driven by the \emph{structural} properties of CoADTP rather than by specific lexical choices, we evaluate three prompts spanning two qualitatively different levels of structure (Table~\ref{tab:Prompt_ablation}):
\begin{itemize}
    \item \textbf{Prompt 1 (Minimal):} A one-line caption instruction (``Describe this scene and generate an anomaly score between 0 and 1'') with no anomaly priors, no reasoning, and no temporal context.
    \item \textbf{Prompt 2 (Format-Structured):} A surveillance-tailored prompt providing \emph{surface} structure bulleted anomaly categories (Violence, Crimes, Disasters, Anomalies) and a defined output format (Event / Type / Details / Confidence) but lacking step-wise reasoning or temporal context.
    \item \textbf{CoADTP (Reasoning-Structured):} The full chain-of-anomaly-detection prompt with explicit step-wise reasoning, a richer anomaly database, and temporal feedback across segments.
\end{itemize}
The gap between Prompt 1 and Prompt 2 is small (+2.92\% AUC, $-0.31$\% AP), indicating that adding \emph{format-level} structure alone categories and output templates yields only marginal gains. In contrast, CoADTP outperforms both (+7.77\% AUC and +16.23\% AP over Prompt 1) by introducing \emph{reasoning-level} structure: step-wise analysis and temporal propagation. Performance scales monotonically with the \emph{depth} of structural complexity rather than surface wording. A component-wise breakdown of which sub-mechanisms (reasoning chain, anomaly database, temporal feedback, cross-modal re-ranking) drive these gains is provided in Table~\ref{table:Prompt_ab} (Appendix~C). This robustness is further confirmed by our cross-backbone results (Sec.~\ref{subsec:quantitative}), where CoADTP improves multiple LVLM backbones (Video-ChatGPT, Video-LLaVA, LLaVA-Next-Video, QwenVL2, InternVL2, Video-R1) by large margins \emph{without} prompt re-tuning, indicating that the prompt structure transfers robustly across model families.

\begin{table}[t]
\centering
\captionsetup{font=normalsize, skip=2pt}
\scriptsize
\footnotesize
\setlength{\tabcolsep}{10pt}
\renewcommand{\arraystretch}{1.00} 
\caption{Ablation of prompting strategies in the RAW stage on UCF-Crime.Prompt~1 is a minimal caption-style prompt; Prompt~2 is a format-structured surveillance prompt with anomaly categories and an output template but no reasoning; CoADTP adds step-wise reasoning, an anomaly database, and temporal feedback.}
\resizebox{1\linewidth}{!}{
\begin{tabular}{lccc|ccc}
\toprule
Prompt & Prior & Reason & Temporal & AUC $\uparrow$ & AP $\uparrow$ & Acc@0.5 $\uparrow$ \\
\midrule
Prompt 1 {\color{blue}(Minimal)} & \xmark & \xmark & \xmark & 0.7111 & 0.2150 & 78.28 \\
Prompt 2 {\color{blue}(Format-Structured)} & \cmark & \xmark & \xmark & 0.7403 & 0.2119 & 57.93 \\
\rowcolor{gray!12}
CoADTP(Raw Stage) & \cmark & \cmark & \cmark & \textbf{0.7888} & \textbf{0.3773} & \textbf{82.41} \\
\bottomrule
\end{tabular}
}
\label{tab:Prompt_ablation}
\end{table}
\subsection{Refinement Stage: Cross-Modal Re-Ranking}
\label{subsec:Analysis}

The raw stage produces independent segment-level anomaly scores $\tilde{\mathbf{A}} = [a_1,\dots,a_h]$, 
which may lack global semantic awareness and long-range temporal consistency. 
To address this limitation, we introduce a parameter-free refinement stage 
that recalibrates each raw score $a_u$ using cross-modal retrieval cues. Let $\mathbf{x}_u$ denote the visual embedding of segment $V_u$. 
The refinement of $a_u$ proceeds in three stages.

\noindent\textbf{1) Global semantic retrieval.} We first retrieve semantically related descriptions 
$\{y_k\}_{k=1}^{M}$ using FAISS-based nearest-neighbor search 
in the shared ImageBind embedding space. The cosine-based text similarity is defined as:
\begin{equation}
S_{\text{text}}(u,k) 
= \cos(\mathbf{x}_u, \mathbf{y}_k),
\label{eq:text_sim}
\end{equation}
where $\mathbf{y}_k$ is the embedding of description $y_k$.

\noindent This step captures global semantic alignment, 
but does not account for temporal proximity within the video.

\noindent\textbf{2) Local visual similarity with temporal weighting.} To enforce temporal coherence, we compute cosine similarity 
between $\mathbf{x}_u$ and all visual embeddings 
$\{\mathbf{x}_t\}$ from the same video:

\begin{equation}
S_{\text{cos}}(u,t) = \cos(\mathbf{x}_u, \mathbf{x}_t).
\end{equation}

We modulate this similarity using a temporal weighting function:
\begin{equation}
w_t =
\begin{cases}
1 + 0.5\, \exp\!\left(-\dfrac{|t-u|}{2\Delta}\right),
& |t-u| \le W\Delta, \\
1, & \text{otherwise},
\end{cases}
\label{eq:temporal_weight}
\end{equation}
where $\Delta$ is the frame interval and $W$ is the temporal window size.

\noindent The temporally-weighted visual similarity becomes:
\begin{equation}
S_{\text{visual}}(u,t)
= S_{\text{cos}}(u,t) \cdot w_t.
\label{eq:visual_sim}
\end{equation}

\noindent This emphasizes visually similar segments that are also 
temporally close to $V_u$, improving local stability.

\noindent\textbf{3) Cross-modal re-ranking.} Each retrieved descriptiopn $y_k$ is associated with a set of 
video frames $\mathcal{F}(k)$. 
For candidate $k$, we compute the maximal visual similarity:

\begin{equation}
\hat{S}_{\text{visual}}(u,k)
= \max_{t \in \mathcal{F}(k)}
S_{\text{visual}}(u,t).
\end{equation}

\noindent We then fuse text and visual similarities:

\begin{equation}
S(u,k)
= \alpha S_{\text{text}}(u,k)
+ (1-\alpha)\hat{S}_{\text{visual}}(u,k),
\quad \alpha \in [0,1].
\label{eq:fused_score}
\end{equation}

\noindent The top-$K$ candidates under $S(u,k)$ 
form the refined neighbor set $\mathcal{K}_u$.

\noindent\textbf{4) Similarity-weighted score aggregation.} The refined anomaly score for segment $u$ is obtained 
by softmax-weighted aggregation of the raw scores 
from its re-ranked neighbors:

\begin{equation}
\hat{a}_u
=
\sum_{k \in \mathcal{K}_u}
a_k
\frac{
\exp\!\left(S(u,k)\right)
}{
\sum_{k' \in \mathcal{K}_u}
\exp\!\left(S(u,k')\right)
}.
\label{eq:refined_score}
\end{equation}

By jointly integrating global semantic alignment 
(Eq.~\ref{eq:text_sim}) and temporally-aware visual consistency 
(Eqs.~\ref{eq:temporal_weight}--\ref{eq:visual_sim}), 
the refinement stage produces smoother, context-aware anomaly scores without introducing additional trainable parameters.

\section{Experimental Setup}
\label{sec:experiments}

In this section, we systematically evaluate Cog-VADU to validate its effectiveness from three complementary perspectives.

\noindent
\textbf{(Q1) Detection Effectiveness.}
Does sequential cognitive prompting with contextual refinement improve zero-shot video anomaly detection?
We evaluate detection performance on large-scale surveillance benchmarks.

\noindent
\textbf{(Q2) Reasoning Depth.}
Does structured anomaly prompting enable deeper semantic understanding beyond binary detection?
We assess reasoning capability on dedicated anomaly-understanding benchmarks.

\noindent
\textbf{(Q3) Generalization.}
How robust is Cog-VADU across datasets with diverse anomaly distributions, annotation protocols, and evaluation criteria?

\paragraph{\textbf{Detection Benchmarks.}}

\textbf{UCF-Crime}~\cite{sultani2019realworldanomalydetectionsurveillance}
contains 13 anomaly categories and 290 test videos (140 abnormal), 
with an average duration of 2.13 minutes. \textbf{XD Violence}~\cite{wu2020looklistenlearningmultimodal}
includes six anomaly types and 800 test videos (500 abnormal),
with an average duration of 1.62 minutes.

{\textbf{Evaluation Metrics.}} Following prior zero-shot VAD works~\cite{zanella2024harnessinglargelanguagemodels,ye2025veraexplainablevideoanomaly},
We report frame-level Area Under the Curve (AUC).
For XD-Violence, we additionally report Average Precision (AP).

\textbf{Reasoning Benchmarks.} \textbf{VANE-Bench}~\cite{bharadwaj2025vanebenchvideoanomalyevaluation}
contains 325 clips and 559 multiple-choice QA pairs across nine sub-datasets, designed to evaluate semantic anomaly understanding.
 \textbf{HIVAU-70k}~\cite{zhang2025holmesvaulongtermvideoanomaly}
evaluates hierarchical anomaly reasoning at clip-, event-, and video-level granularity.
Annotations are constructed via semi-automated segmentation and recursive description.

{\textbf{Evaluation Metrics.}} 
To evaluate semantic anomaly understanding,
performance on {VANE-Bench}~\cite{bharadwaj2025vanebenchvideoanomalyevaluation}, we measured MCQ accuracy . For {HIVAU-70k}~\cite{zhang2025holmesvaulongtermvideoanomaly}, we report BLEU, METEOR, and ROUGE-L.
Together, these benchmarks evaluate detection accuracy, structured reasoning ability, and cross-dataset generalization under a fully training-free setting.
\subsection{Benchmarks and Unified Re-annotation}
\label{subsec:reanno}
\begin{figure}[htbp]
\centering
\captionsetup{font=normalsize, skip=5pt}
\includegraphics[width=1.0\textwidth]{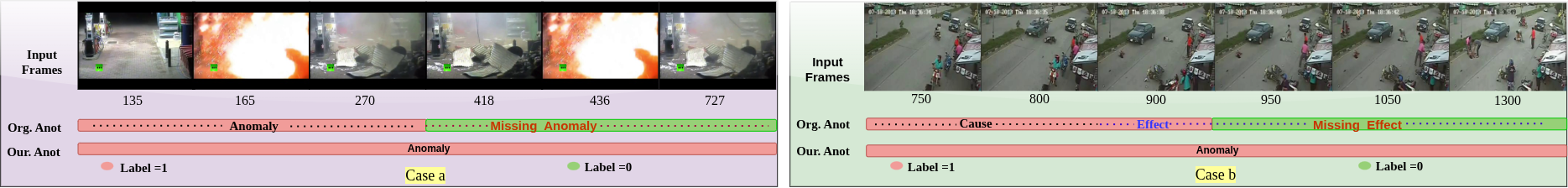}
\caption{\textbf{UCF~\cite{sultani2019realworldanomalydetectionsurveillance} Re-Annotation Examples.}  
Case (a) shows missing anomaly regions in the original annotations.  
Case (b) highlights cause–and–effect anomaly segments previously ignored but included in our unified annotations.}
\label{fig:reanot}
\end{figure}

\noindent \textbf{Unified Re-annotation of Cause-Effect Relationships.}
While evaluating on UCF-Crime~\cite{sultani2019realworldanomalydetectionsurveillance} and XD-Violence~\cite{wu2020looklistenlearningmultimodal}, we identified systematic annotation inconsistencies that penalize reasoning-centric models. Specifically, we observed a ``trigger-bias'' where only the initial cause of an anomaly (e.g., a gunshot) is labeled, while semantically anomalous effects (e.g., victims collapsing, smoke, or aftermath) are omitted. Additionally, repetitive anomalies are often inconsistently marked, frequently appearing at $t_n$ without labels despite a labeled occurrence at $t_1$. As illustrated in Fig.~\ref{fig:reanot}, these gaps create a structural mismatch between visual reasoning and ground-truth labels. To address this, we re-annotated both datasets using a unified \emph{causal-chain policy} that captures the complete temporal extent of each event-from the initiating cause to the resulting effect and escalation. This protocol increased anomalous frame counts by 39.1\% and 22.5\% for UCF-Crime and XD-Violence, respectively. \textit{Detailed statistics and the full annotation protocol are provided in the Supplementary Material (Sec:\ref{sec:ANOT_protocol}}).

\noindent \textbf{Implementation of Cog-VADU.}
We adopt \textit{VideoLLaMA3-7B}~\cite{zhang2025videollama3frontiermultimodal} as the default backbone $f_{\mathrm{LVLM}}$ for Cog-VADU, and report all results under this setting unless otherwise specified. For ablation studies, we additionally evaluate alternative backbones, including \textit{VideoLLaMA3-2B}~\cite{zhang2025videollama3frontiermultimodal} and \textit{InternVL2-8B}~\cite{chen2024internvlscalingvisionfoundation}. During the refinement stage, we use \textit{ImageBind}~\cite{girdhar2023imagebindembeddingspacebind} to extract cross-modal retrieval features. We select $M=30$ top visual–textual candidates, followed by softmax-weighted aggregation over the top $k=10$ reranked candidates. In the raw stage, we sample $N=10$ frames per temporal segment of length $T=10$ seconds, with a frame interval $\Delta=16$.

\subsection*{Quantitative Results}
\label{subsec:quantitative}
We compare Cog-VADU with four representative VAD paradigms:
(1) weakly supervised, (2) one-class unsupervised, (3) training-based explainable, and (4) training-free explainable methods. Results are reported in Table~\ref{tab:xdviolence_auc} and Table~\ref{tab:ucfcrime_auc}.

\noindent \textbf{Detection Performance.}
Despite being fully training-free, Cog-VADU achieves a competitive AUC-ROC across both datasets. On XD-Violence, it reaches 90.15\% AUC, surpassing the previous training-free state-of-the-art, LAVAD~\cite{zanella2024harnessinglargelanguagemodels} (85.36\%), and outperforming several training-based explainable approaches. This demonstrates that structured anomaly reasoning alone enables strong temporal discrimination without dataset-specific optimization.

\noindent \textbf{AP under Original Labels.}
Although AUC remains high, Average Precision (AP) under the original annotations is lower than that of supervised models. As discussed in Section~\ref{subsec:reanno}, this is due to the \emph{trigger bias}, where only initiating causes are labeled while downstream effects are omitted. Supervised models implicitly learn to ignore these regions, whereas Cog-VADU assigns high anomaly scores to both causes and consequences. Consequently, correct causal reasoning may be penalized as false positives under an incomplete ground truth.

\noindent \textbf{Performance under Unified Re-annotation.}
Under our unified re-annotation protocol (“R.”), the precision gap largely disappears. This recovery confirms that Cog-VADU captures coherent cause–and–effect chains rather than isolated triggers and motion patterns, thus aligning more closely with human semantic understanding of anomalies.


\begin{table}[htbp]
\centering
\small
\captionsetup{font=normalsize, skip=2pt}
\begin{minipage}[t]{0.493\textwidth}
\centering
\captionsetup{font=normalsize, skip=2pt}
\setlength{\tabcolsep}{0.0001pt}
\renewcommand{\arraystretch}{1}
\caption{\textbf{AUC and AP (\%) on XD-Violence}~\cite{wu2020looklistenlearningmultimodal}.
``O.'' and ``R.'' denote original and re-annotated splits.
``---'' indicates codes/results unavailable.}
\label{tab:xdviolence_auc}
\resizebox{\textwidth}{!}{%
\begin{tabular}{lcccc}
\toprule
\textbf{Method} & \textbf{AUC (O.)} & \textbf{AUC (R.)} & \textbf{AP (O.)} & \textbf{AP (R.)} \\
\midrule
\rowcolor{NavyBlue!20}
\multicolumn{5}{l}{\textbf{Weakly Supervised (Non-Explainable)}} \\
\rowcolor{NavyBlue!5}
RareAnom~\cite{10.1016/j.patcog.2023.109567}                    & 68.33          & ---            & ---            & ---   \\
\rowcolor{NavyBlue!10}
RTFM~\cite{tian2021weaklysupervisedvideoanomalydetection}        & ---            & ---            & 77.81          & ---   \\
\rowcolor{NavyBlue!5}
MGFN~\cite{chen2022mgfnmagnitudecontrastiveglanceandfocusnetwork}& ---            & ---            & 80.11          & ---   \\
\rowcolor{NavyBlue!10}
UR-DMU~\cite{zhou2023dualmemoryunitsuncertainty}                 & \textbf{94.02} & \textbf{93.74} & \textbf{81.66} & \textbf{84.50} \\
\midrule
\rowcolor{MidPurple!20}
\multicolumn{5}{l}{\textbf{One-Class (Non-Explainable)}} \\
\rowcolor{MidPurple!10}
BODS~\cite{wang2019godsgeneralizedoneclassdiscriminative}        & 57.32          & ---            & ---            & ---   \\
\rowcolor{MidPurple!5}
GODS~\cite{wang2019godsgeneralizedoneclassdiscriminative}        & 61.56          & ---            & ---            & ---   \\
\midrule
\rowcolor{ForestGreen!25}
\multicolumn{5}{l}{\textbf{Training-Based (Explainable)}} \\
\rowcolor{ForestGreen!10}
VERA~\cite{ye2025veraexplainablevideoanomaly}                    & 88.26          & 92.15          & 70.54          & 80.62 \\
\rowcolor{ForestGreen!5}
Holmes-VAU~\cite{zhang2025holmesvaulongtermvideoanomaly}         & ---            & ---            & \textbf{87.68} & ---   \\
\midrule
\rowcolor{BurntOrange!25}
\multicolumn{5}{l}{\textbf{Training-Free (Explainable)}} \\
\rowcolor{BurntOrange!10}
ZS CLIP~\cite{zanella2024harnessinglargelanguagemodels}          & 38.21          & ---            & 17.83          & ---   \\
\rowcolor{BurntOrange!5}
ZS IB (Img)~\cite{girdhar2023imagebindembeddingspacebind}        & 27.25          & ---            & ---            & ---   \\
\rowcolor{BurntOrange!10}
ZS IB (Vid)~\cite{girdhar2023imagebindembeddingspacebind}        & 25.36          & ---            & ---            & ---   \\
\rowcolor{BurntOrange!5}
LLAVA-1.5~\cite{liu2024improvedbaselinesvisualinstruction}       & 79.62          & ---            & 50.26          & ---   \\
\rowcolor{BurntOrange!10}
LAVAD~\cite{zanella2024harnessinglargelanguagemodels}            & 85.36          & 88.72          & 62.01          & 75.44 \\
\rowcolor{BurntOrange!5}
\textbf{Ours}                                                    & \textbf{90.15} & \textbf{94.32} & \textbf{65.15} & \textbf{95.24} \\
\bottomrule
\end{tabular}}

\end{minipage}
\hfill
\begin{minipage}[t]{0.475\textwidth}
\centering
\setlength{\tabcolsep}{10pt}
\renewcommand{\arraystretch}{0.958}
\caption{\textbf{AUC (\%) on UCF-Crime}~\cite{sultani2019realworldanomalydetectionsurveillance}.
``O.'' and ``R.'' denote original and re-annotated splits.}
\label{tab:ucfcrime_auc}
\resizebox{\textwidth}{!}{%
\begin{tabular}{lcc}
\toprule
\textbf{Method} & \textbf{AUC (O.)} & \textbf{AUC (R.)} \\
\midrule
\rowcolor{NavyBlue!20}
\multicolumn{3}{l}{\textbf{Weakly Supervised (Non-Explainable)}} \\
\rowcolor{NavyBlue!5}
CLAWS~\cite{zaheer2021clawsclusteringassistedweakly}                                                         & 83.03          & ---   \\
\rowcolor{NavyBlue!10}
RTFM~\cite{tian2021weaklysupervisedvideoanomalydetection}        & 84.03          & ---   \\
\rowcolor{NavyBlue!5}
UR-DMU~\cite{zhou2023dualmemoryunitsuncertainty}                 & 86.97          & 89.57 \\
\rowcolor{NavyBlue!10}
MGFN~\cite{chen2022mgfnmagnitudecontrastiveglanceandfocusnetwork}& 86.98          & ---   \\
\rowcolor{NavyBlue!5}
SSRL~\cite{li2022scale}                                         & 87.43          & ---   \\
\rowcolor{NavyBlue!10}
CLIP-TSA~\cite{joo2023clip}                                     & \textbf{87.58} & ---   \\
\midrule
\rowcolor{MidPurple!20}
\multicolumn{3}{l}{\textbf{One-Class (Non-Explainable)}} \\
\rowcolor{MidPurple!10}
BODS~\cite{wang2019godsgeneralizedoneclassdiscriminative}        & 68.26          & ---   \\
\rowcolor{MidPurple!5}
GODS~\cite{wang2019godsgeneralizedoneclassdiscriminative}        & \textbf{70.46} & ---   \\
\midrule
\rowcolor{ForestGreen!25}
\multicolumn{3}{l}{\textbf{Training-Based (Explainable)}} \\
\rowcolor{ForestGreen!10}
VERA~\cite{ye2025veraexplainablevideoanomaly}                    & 86.55          & 88.34 \\
\rowcolor{ForestGreen!5}
Holmes-VAU~\cite{zhang2025holmesvaulongtermvideoanomaly}         & \textbf{88.96} & \textbf{89.52} \\
\midrule
\rowcolor{BurntOrange!25}
\multicolumn{3}{l}{\textbf{Training-Free (Explainable)}} \\
\rowcolor{BurntOrange!10}
ZS CLIP~\cite{zanella2024harnessinglargelanguagemodels}          & 53.16          & ---   \\
\rowcolor{BurntOrange!5}
ZS IB (Img)~\cite{girdhar2023imagebindembeddingspacebind}        & 53.65          & ---   \\
\rowcolor{BurntOrange!10}
ZS IB (Vid)~\cite{girdhar2023imagebindembeddingspacebind}        & 55.78          & ---   \\
\rowcolor{BurntOrange!5}
LLAVA-1.5~\cite{liu2024improvedbaselinesvisualinstruction}       & 72.84          & ---   \\
\rowcolor{BurntOrange!10}
LAVAD~\cite{zanella2024harnessinglargelanguagemodels}            & 80.28          & 83.80 \\
\rowcolor{BurntOrange!5}
\textbf{Ours}                                                    & \textbf{80.54} & \textbf{86.06} \\
\bottomrule
\end{tabular}}
\end{minipage}

\end{table}

\subsection{Anomaly Understanding \& Reasoning (Q2): Model-Agnostic Evaluation}

We evaluate hierarchical anomaly reasoning on HIVAU-70K across clip-level (C), event-level (E), and video-level (V) granularities. 
Table~\ref{tab:reasoning_comparison} compares three groups: 
(1) training + instruction-tuned VAD MLLMs, 
(2) open-source MLLMs, and 
(3) open-source MLLMs augmented with CoADTP. 
Cog-VADU represents our full training-free framework.

\noindent \textbf{Baseline MLLMs Lack Structured Anomaly Reasoning.}
General open-source MLLMs (e.g., Video-ChatGPT, Video-LLaVA, LLaVA-Next-Video, InternVL2) exhibit weak event- and video-level performance. 
Although some models achieve moderate clip-level BLEU, their scores collapse at longer temporal scales, indicating insufficient anomaly grounding and poor hierarchical reasoning.

\noindent \textbf{CoADTP is Model-Agnostic and Consistently Improves Anomaly Reasoning.}
Adding CoADTP to the same backbones yields dramatic gains across all granularities. 
These consistent improvements across diverse architectures demonstrate that CoADTP is model-agnostic: it enhances anomaly reasoning without modifying model weights or requiring task-specific training. 
Structured chain-of-anomaly prompting alone substantially strengthens temporal coherence and causal understanding.

\noindent \textbf{Comparison with Training-Based VAD MLLMs.}
Training-based models such as Hawk \cite{tang2024hawklearningunderstandopenworld}, Holmes-VAD, \cite{zhang2024holmesvadunbiasedexplainablevideo}, and Holmes-VAU \cite{zhang2025holmesvaulongtermvideoanomaly} benefit from anomaly-focused supervision and hierarchical instruction tuning, with Holmes-VAU achieving the strongest clip-level BLEU and ROUGE-L scores due to dataset-specific optimization. In contrast, enforcing cognitive reasoning through CoADTP yields consistent improvements across multiple backbones, demonstrating that CoADTP is a robust, model-agnostic framework rather than a backbone-dependent enhancement.

\begin{table}[htbp]
\centering
\captionsetup{font=normalsize, skip=4pt}
\caption{
Comparison of hierarchical video anomaly reasoning performance across clip-level (C), event-level (E), and video-level (V) granularity.
BLEU denotes cumulative BLEU-1 to BLEU-4 scores.
Results demonstrate the effectiveness of reasoning-driven approaches for structured anomaly understanding from short clips to long videos. \textbf{Bold} means best and \underline{Underline} means second best in \textbf{each group}.
Results with + CoADTP show the impact of the Chain of Anomaly Detection Thought Prompt.
}
\label{tab:reasoning_comparison}
\footnotesize
\setlength{\tabcolsep}{3pt}  
\renewcommand{\arraystretch}{0.95} 
\resizebox{\textwidth}{!}{
\begin{tabular}{l |ccc| |ccc| |ccc}
\toprule
\rowcolor{gray!15} \multirow{2}{*}{Method}
& \multicolumn{3}{c}{BLEU $\uparrow$}
& \multicolumn{3}{c}{METEOR $\uparrow$}
& \multicolumn{3}{c}{ROUGE-L $\uparrow$} \\
\cmidrule(lr){2-4} \cmidrule(lr){5-7} \cmidrule(lr){8-10}
& C & E & V & C & E & V & C & E & V \\
\midrule
\rowcolor{gray!15} \multicolumn{10}{l}{\textbf{Training + Instruction-Tuned VAD MLLMs}} \\
\midrule
Hawk ~\cite{tang2024hawklearningunderstandopenworld}
& 0.320 & 0.165 & 0.202
& \textbf{0.228} & 0.191 & 0.196
& 0.156 & 0.104 & 0.114 \\
Holmes-VAD ~\cite{zhang2024holmesvadunbiasedexplainablevideo}
& \textit{0.514} & \textit{0.318} & \textit{0.306}
& \textit{0.224} & \textbf{0.237} & \textbf{0.211}
&\textit{0.235} & \textit{0.164} & \textit{0.161} \\
Holmes-VAU ~\cite{zhang2025holmesvaulongtermvideoanomaly}
& \textbf{0.913} & \textbf{0.804} & \textbf{0.566}
& 0.190 & 0.165 & 0.121
& \textbf{0.329} & \textbf{0.370} & \textbf{0.355} \\
\hline
\rowcolor{gray!15} \multicolumn{10}{l}{\textbf{Open-source MLLMs}} \\
\midrule
Video-ChatGPT ~\cite{maaz2024videochatgptdetailedvideounderstanding}
& 0.152 & 0.068 & 0.066
& 0.102 & 0.069 & 0.044
& 0.153 & 0.048 & 0.079 \\
\rowcolor{gray!30}Video-ChatGPT + CoADTP(Ours)
& 0.349 & 0.439 & 0.447
& 0.197 & 0.253 & 0.242
& 0.162 & 0.222 & 0.234 \\
Video-LLaVA ~\cite{lin2024videollavalearningunitedvisual}
& 0.164 & 0.046 & 0.055
& 0.097 & 0.022 & 0.014
& 0.132 & 0.023 & 0.045 \\
\rowcolor{gray!30}Video-LLaVA + CoADTP(Ours)
& 0.391 & 0.213 & 0.230
& 0.256 & 0.209 & 0.199
& 0.183 & 0.131 & 0.137 \\
LLaVA-Next-Video~\cite{zhang2024llavanext-video}
& 0.435 & 0.091 & 0.120
& 0.117 & 0.085 & 0.096
& 0.198 & 0.080 & 0.106 \\
\rowcolor{gray!30}LLaVA-Next-Video + CoADTP(Ours)
& \underline{0.651} & \textbf{0.839} & \textbf{0.790}
& 0.287 & \textbf{0.339} & \underline{0.299}
&  {0.253} &  \underline{0.352} &  \underline{0.325} \\
QwenVL2~\cite{wang2024qwen2vlenhancingvisionlanguagemodels}
& 0.312 & 0.082 & 0.155
& 0.133 & 0.092 & 0.112
& 0.163 & 0.081 & 0.137 \\
\rowcolor{gray!30}QwenVL2 + CoADTP(Ours)
& 0.470 & {0.792} & 0.722
& 0.215 & 0.307 & 0.281
& 0.237 & 0.341 & 0.323 \\
InternVL2~\cite{chen2024internvl}
& 0.331 & 0.101 & 0.145
& 0.141 & 0.095 & 0.101
& 0.182 & 0.102 & 0.122 \\
\rowcolor{gray!30}InternVL2 + CoADTP(Ours)
& 0.550 &  \underline{0.830} & \underline{0.766}
& 0.247 & \underline{0.323} &  \textbf{0.299}
& \textbf{0.281} &  \textbf{0.357} &  \textbf{0.342} \\

Video-R1 ~\cite{feng2025video}
& 0.572 & 0.171 & 0.229
& 0.186 & 0.159 & 0.156
&  0.187 & 0.103 & 0.116 \\
\rowcolor{gray!30}Video-R1 + CoADTP(Ours)
& 0.439 & 0.769 &0.732
& \underline{0.338} & 0.301&  0.280
& 0.236 & 0.347 &  0.325  \\

\rowcolor{gray!15}{Cog-VADU (Ours)}
& \textbf{0.785} & {0.739} & {0.743}
& \textbf{0.349} & {0.297} &{0.290}
& \underline{0.278} & {0.329} & {0.324} \\
\bottomrule
\end{tabular}
`}
\end{table}

\paragraph{Beyond n-gram metrics: semantic faithfulness via BERTScore.}
The metrics in Table~\ref{tab:reasoning_comparison}%

quantify lexical overlap with reference rationales, which can penalise 
valid paraphrases and reward surface-level word matches. To complement 
the n-gram evaluation, we additionally report 
BERTScore~\cite{zhang2020bertscoreevaluatingtextgeneration}, which 
measures semantic similarity via contextual embeddings of 
RoBERTa-large~\cite{zhang2020bertscoreevaluatingtextgeneration} with 
baseline-rescaling. Whereas Table~\ref{tab:reasoning_comparison} 
partitions results by temporal granularity (clip, event, video), 
Table~\ref{tab:bertscore_main} partitions by HIVAU-70k task type 
(judgement, description, analysis, captioning), giving a complementary 
view of rationale quality across the four reasoning instructions the 
benchmark prescribes. We report F1 on the intersection of 350 IDs 
covered by every evaluated method.

\begin{table}[!ht]
\centering
\captionsetup{font=small, skip=4pt}
\caption{BERTScore F1 \cite{zhang2020bertscoreevaluatingtextgeneration}on HIVAU-70k~\cite{zhang2025holmesvaulongtermvideoanomaly} 
(intersection of 350 IDs covered by all 10 methods; 
RoBERTa-large~\cite{zhang2020bertscoreevaluatingtextgeneration}, 
baseline-rescaled). Cog-VADU achieves training-free performance 
competitive with strong CoADTP-augmented backbones while 
substantially outperforming the training-based explainable baseline 
Holmes-VAD~\cite{zhang2024holmesvadunbiasedexplainablevideo}. 
Best in each column is in \textbf{bold}; second-best is 
\underline{underlined}.}
\label{tab:bertscore_main}
\small
\setlength{\tabcolsep}{6pt}
\renewcommand{\arraystretch}{1.2}
\begin{tabular}{l|c|c|c|c|c}
\toprule
\textbf{Method} & \textbf{Judg.} & \textbf{Desc.} & \textbf{Anal.} & \textbf{Capt.} & \textbf{Overall} \\
\midrule
\multicolumn{6}{l}{\textit{Training + instruction-tuned VAD MLLMs}} \\
HAWK~\cite{tang2024hawklearningunderstandopenworld}                                  & 0.078 & 0.075 & 0.071 & 0.077 & 0.077 \\
Holmes-VAD~\cite{zhang2024holmesvadunbiasedexplainablevideo}                         & 0.167 & 0.236 & 0.235 & 0.263 & 0.245 \\
Holmes-VAU~\cite{zhang2025holmesvaulongtermvideoanomaly} (SOTA, in-domain)           & \textbf{0.652} & \textbf{0.297} & \textbf{0.479} & \textbf{0.430} & \textbf{0.431} \\
\midrule
\multicolumn{6}{l}{\textit{Open-source MLLMs + CoADTP (zero-shot)}} \\
Video-ChatGPT~\cite{maaz2024videochatgptdetailedvideounderstanding} + CoADTP           & 0.400 & 0.170 & 0.084 & 0.104 & 0.149 \\
Video-LLaVA~\cite{lin2024videollavalearningunitedvisual} + CoADTP                    & 0.114 & 0.081 & 0.129 & 0.143 & 0.127 \\
LLaVA-Next-Video~\cite{zhang2024llavanext-video} + CoADTP                             & 0.555 & 0.197 & 0.243 & 0.333 & 0.326 \\
QwenVL2~\cite{wang2024qwen2vlenhancingvisionlanguagemodels} + CoADTP                 & 0.507 & 0.237 & 0.259 & 0.333 & 0.330 \\
InternVL2~\cite{chen2024internvl} + CoADTP                                           & 0.562 & 0.235 & 0.277 & 0.355 & \underline{0.350} \\
Video-R1~\cite{feng2025video} + CoADTP                                               & 0.516 & 0.224 & 0.288 & 0.269 & 0.288 \\
\midrule
\rowcolor{green!15}
\textbf{Cog-VADU (Ours, VideoLLaMA3~\cite{zhang2025videollama3frontiermultimodal})}  & \underline{0.545} & 0.223 & 0.264 & 0.309 & 0.316 \\
\bottomrule
\end{tabular}
\end{table}

\paragraph{Key findings.}
Four observations follow from Table~\ref{tab:bertscore_main}. 
\emph{First}, Cog-VADU attains an overall F1 of $0.316$ 
\emph{without any task-specific training}, substantially outperforming 
the training-based explainable baseline 
Holmes-VAD~\cite{zhang2024holmesvadunbiasedexplainablevideo} ($0.245$) 
and HAWK~\cite{tang2024hawklearningunderstandopenworld} ($0.077$). 
\emph{Second}, on the \emph{judgement} task --- which directly tests 
the model's anomaly decision, Cog-VADU achieves F1 $= 0.545$, the 
second-highest among all methods and close to Holmes-VAU's $0.652$ 
despite never being trained on HIVAU-70k. 
\emph{Third}, the gains from CoADTP scale with backbone capability: 
stronger backbones (InternVL2, QwenVL2, LLaVA-Next-Video, Video-R1) 
cluster in the $0.29$--$0.35$ range, while older backbones with 
weaker instruction-following (Video-ChatGPT, Video-LLaVA) remain at 
$0.13$--$0.15$. This pattern is consistent with the cross-backbone 
trends in Table~\ref{tab:reasoning_comparison} indicates 
that CoADTP \emph{amplifies the reasoning capacity already present in 
the underlying LVLM} rather than substituting for it. 
\emph{Fourth}, the gap to Holmes-VAU~\cite{zhang2025holmesvaulongtermvideoanomaly} 
($0.431$) is attributable to its in-domain instruction tuning on 
HIVAU-70k; given that Cog-VADU is fully training-free and incurs no 
annotation cost, this trade-off is expected and acceptable.

Taken together, BERTScore confirms that Cog-VADU's predictions are 
\emph{semantically faithful} to the references even when phrasing 
differs from the ground truth, complementing rather than replacing 
the n-gram metrics in Table~\ref{tab:reasoning_comparison}.

\subsection{Performance Comparison on VANE-Bench \& Generalization (Q3)}

As shown in Table~\ref{tab:vane_comps}, we evaluate our zero-shot, reasoning-based VAD framework Cog-VADU on VANE-Bench~\cite{bharadwaj2025vanebenchvideoanomalyevaluation}, a challenging benchmark designed to assess video anomaly understanding through multi-type, multiple-choice questions. Results are reported in terms of {anomaly detection accuracy}. Despite not relying on any task-specific training or instruction tuning, our \textbf{Cog-VADU} achieves competitive performance against strong baselines, demonstrating the effectiveness of structured cognitive reasoning for anomaly detection.
\begin{table}[H]
\centering
\captionsetup{font=normalsize, skip=4pt}
\scriptsize
\setlength{\tabcolsep}{2pt}
\renewcommand{\arraystretch}{1.0}

\caption{\textbf{Performance comparison on VANE-Bench.} \cite{bharadwaj2025vanebenchvideoanomalyevaluation}
Accuracy (\%) across different video generation models and surveillance datasets. 
\textbf{bold} means best and \underline{underline} means second best in a group.}
\label{tab:vane_comps}

\begin{tabular}{l|c|c|c|c|c|c|c|c}
\toprule
\textbf{Method} & \textbf{SORA} & \textbf{OpenSORA} & \textbf{RG2} & \textbf{VideoLCM} & \textbf{MS-T2} & \textbf{Avenue} & \textbf{Ped1} & \textbf{Ped2} \\

\midrule
\rowcolor{ForestGreen!20}
\multicolumn{9}{l}{\textbf{MLLM-based VAD (Training+ instruction Tuned )}} \\
\rowcolor{ForestGreen!10}
Holmes-VAU~\cite{zhang2025holmesvaulongtermvideoanomaly}         & 2.17  & 34.00 & 24.00 & 29.81 & 25.00 & 6.06  & 3.33  & 5.56 \\
\rowcolor{ForestGreen!5}
Holmes-VAD~\cite{zhang2024holmesvadunbiasedexplainablevideo}         & 6.52  & 34.00 & 32.00 & 33.56 & 22.92 & 12.12 & 20.00 & 5.56 \\
\rowcolor{ForestGreen!10}
HAWK~\cite{tang2024hawklearningunderstandopenworld}                   & \underline{24.64} & \underline{52.00} & \underline{44.00} & \underline{36.54} &\underline{50.00} & \underline{36.36} & \underline{36.67} & \underline{38.89} \\
\rowcolor{ForestGreen!5}
    VAD-R1~\cite{huang2025vadr1videoanomalyreasoning}                   & \textbf{41.30} & \textbf{78.00} & \textbf{56.00} & \textbf{63.46} & \textbf{60.42} & \textbf{75.76} & \textbf{60.00} & \textbf{63.89} \\

\midrule
\rowcolor{NavyBlue!15}
\multicolumn{9}{l}{\textbf{Open-Source MLLMs }} \\
\rowcolor{NavyBlue!5}
Video-LLaMA~\cite{zhang2023videollamainstructiontunedaudiovisuallanguage}      & 11.59 & 18.00 & 16.00 & 10.57 & 10.41 & 30.00 & 16.66 & 5.55 \\
\rowcolor{NavyBlue!10}
VideoChat~\cite{li2024videochatchatcentricvideounderstanding}         & 10.74 &\underline{28.00}& 4.00  & 17.64 & 20.83 & 32.25 & 13.33 & 13.88 \\
\rowcolor{NavyBlue!5}
Video-ChatGPT~\cite{maaz2024videochatgptdetailedvideounderstanding}  & \textbf{26.47} & 22.00 & 12.00 & 18.26 & 16.66 & \underline {39.39} & \underline{40.00} & \underline{19.44} \\
\rowcolor{NavyBlue!10}
Video-LLaVA~\cite{lin2024videollavalearningunitedvisual}      & 10.86 & 18.00 & 16.00 & 19.23 & 16.66 & 3.03  & 2.77  & 6.06 \\
\rowcolor{NavyBlue!5}
MovieChat~\cite{song2024moviechatdensetokensparse}         & 8.69  & 10.00 & 16.00 & 14.42 & 6.25  & 18.18 & 6.66  & 11.11 \\
\rowcolor{NavyBlue!10}
LLaMA-VID~\cite{li2023llamavidimageworth2}          & 7.97  & 14.00 & 20.00 & 19.23 & 14.58 & 27.27 & 6.66  & 19.44 \\
\rowcolor{NavyBlue!5}
TimeChat~\cite{ren2024timechattimesensitivemultimodallarge}           & {21.73} & { 26.00} & \underline{28.00 }& \underline{22.11} & \underline{20.83} & {24.20} & {27.58} & {11.11} \\

\midrule
\rowcolor{BurntOrange!20}
\textbf{Cog-VADU(Reasoning-Based Training-Free Ours)}   & \underline{24.64} & \textbf{80.00} & \textbf{60.00} & \textbf{50.00} & \textbf{50.00} & \textbf{72.73} & \textbf{56.67} & \textbf{72.22%
} \\
\bottomrule
\end{tabular}
\end{table}
Importantly, while most existing methods depend on supervised fine-tuning or instruction-tuned representations, Cog-VADU operates in a purely {training-free, zero-shot} setting. This highlights its strong {generalization capability}, as it can transfer to diverse anomaly scenarios without exposure to domain-specific annotations. The results suggest that explicitly modeling reasoning steps, rather than relying solely on learned representations, can serve as a powerful alternative for video anomaly understanding.

\begin{figure}[htbp] 
\centering
\captionsetup{font=normalsize, skip=4pt}
\includegraphics[width=1.0\textwidth]{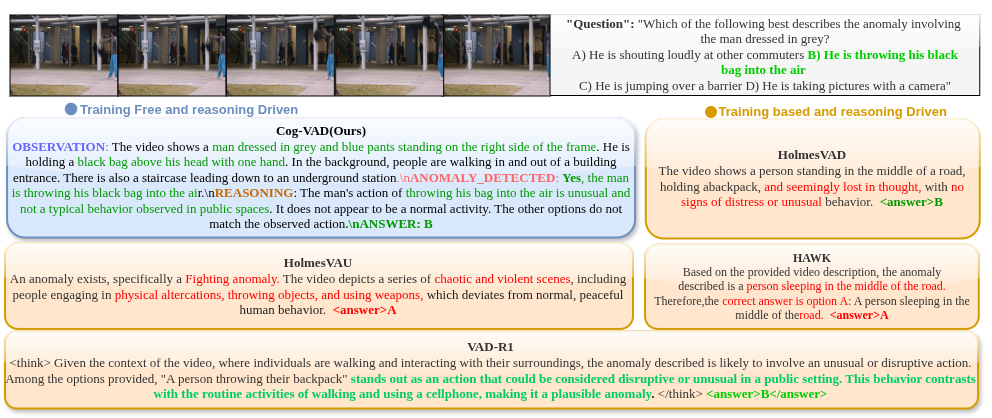}
\caption{\textbf{Qualitative comparison on VANE.} Cog-VADU generates structured, evidence-grounded reasoning (Observation $\rightarrow$ Anomaly $\rightarrow$ Reasoning $\rightarrow$ Answer), correctly identifying the anomaly (B). In contrast, prior multimodal VAD methods, including RL-based VAD-R1~\cite{bharadwaj2025vanebenchvideoanomalyevaluation}, often produce weakly grounded explanations.}
\label{fig:vaneq}
\end{figure}

\noindent \textbf{Qualitative comparison on VANE.} Fig.~\ref{fig:vaneq} further present qualitative examples from VANE-Bench. Our method generates structured and interpretable outputs that explicitly connect \emph{scene understanding} $\rightarrow$ \emph{abnormal event localization} $\rightarrow$ \emph{causal reasoning} $\rightarrow$ \emph{final decision}. Compared to baseline methods, which often produce shallow or implicit predictions, Cog-VADU provides {grounded, step-by-step explanations} that enhance both interpretability and reliability. This demonstrates that our approach not only predicts anomalies effectively but also explains \emph{why} they occur, aligning with the goal of trustworthy video understanding systems.
\subsection{Ablation Studies}
\label{sec:ablation}

We conduct comprehensive ablation studies on the UCF-Crime dataset~\cite{sultani2019realworldanomalydetectionsurveillance} to systematically analyze the contribution of each component in our framework. In particular, we investigate how different design choices affect the quality, stability, and reliability of anomaly predictions. 

Specifically, we examine:  
(1) alternative refinement strategies to understand the role of cross-modal alignment,  
(2) the impact of top-$K$ semantic neighbor selection on contextual aggregation,  
(3) the influence of the visual-textual fusion weight $\alpha$ on balancing global semantic and local temporal cues, and  
(4) the choice of vision-language backbone $f_{\text{LVLM}}$ to evaluate model generalization.  

Through these analyses, we aim to answer three key questions:  
\emph{(i)} how much each component contributes to performance gains,  
\emph{(ii)} how sensitive the framework is to parameter variations, and  
\emph{(iii)} whether the proposed design leads to stable and consistent predictions across diverse settings.  

\noindent\begin{minipage}[htbp]{0.52\textwidth}

\noindent\textbf{Effect of refinement strategies.}
Table~\ref{tab:refinement_strategies_TAB} compares different refinement mechanisms. 
Our \emph{Cross-Modal Re-Ranking} consistently outperforms both visual-textual and visual-visual alternatives, achieving the highest AUC and AP. 

Visual-visual matching primarily captures local temporal consistency, but often fails to incorporate global semantic context. 
In contrast, visual-textual matching emphasizes global semantic alignment, while lacking sensitivity to local temporal coherence. 
Our approach integrates both perspectives, jointly modeling global semantics and local temporal relationships, leading to more reliable and robust anomaly scoring.

\end{minipage}%
\hfill
\begin{minipage}[htbp]{0.45\textwidth}
\centering
\footnotesize
\setlength{\tabcolsep}{5pt}
\renewcommand{\arraystretch}{1.6}
\resizebox{\textwidth}{!}{%
\begin{tabular}{l|c|c}
\toprule
\textbf{Strategy} & \textbf{AUC} & \textbf{AP} \\
\midrule
Visual–Textual~\cite{zanella2024harnessinglargelanguagemodels} & 0.722 & 0.211 \\
Visual–Visual~\cite{ye2025veraexplainablevideoanomaly} & 0.755 & 0.249 \\
\rowcolor{lightgray!25}\textbf{Ours (Cross-Modal Re-Ranking)} & \textbf{0.805} & \textbf{0.381} \\
\bottomrule
\end{tabular}%
}

\captionsetup{font=scriptsize, skip=6pt}
\captionof{table}{\textbf{Ablation: Refinement Strategies (UCF-Crime~\cite{sultani2019realworldanomalydetectionsurveillance}).} Frame-level AUC and AP for different strategies.}
\label{tab:refinement_strategies_TAB}

\end{minipage}

\noindent\textbf{Effect of Refinement Parameters.}
We further analyze the impact of key parameters that govern the refinement stage, including the number of retrieved candidates ($M$), the temporal aggregation window ($K$), and the fusion weights ($\alpha_{\text{vis}}, \alpha_{\text{text}}$). These parameters control semantic retrieval, temporal smoothing, and cross-modal alignment, which are critical for producing stable and reliable anomaly predictions.As shown in Table~\ref{tab:ablation_combined}, increasing $M$ improves performance

\vspace{4pt}

\noindent
\begin{minipage}[htbp]{0.505\textwidth}
\small
\textbf{Analysis.}

\begin{itemize}
\setlength{\itemsep}{4pt}

\item \textbf{Top-$M$:}
Increasing the number of retrieved semantic candidates improves performance by providing richer contextual evidence. Small $M$ leads to unstable alignment, while larger $M$ captures diverse context. However, very large $M$ may introduce noise. Empirically, $M=30$ achieves the best trade-off.

\item \textbf{Temporal window $K$:}
Larger $K$ improves temporal consistency by smoothing predictions and capturing cause-and-effect relations. However, excessive smoothing may blur short anomalies. Optimal performance is observed for $K \in [8,10]$.

\item \textbf{Fusion weights:}
The fusion weights balance the contribution of visual similarity and semantic alignment. 
Visual similarity captures local temporal coherence, while textual similarity provides global semantic context. 
An imbalanced weighting biases the model toward either local or global cues, leading to suboptimal performance. 
In contrast, balanced fusion ($\alpha_{\text{vis}} = \alpha_{\text{text}} = 0.5$) effectively integrates both signals, resulting in stable and context-aware anomaly predictions.
\end{itemize}

\end{minipage}
\hfill
\begin{minipage}[htbp]{0.47\textwidth}

\centering
\includegraphics[width=\textwidth]{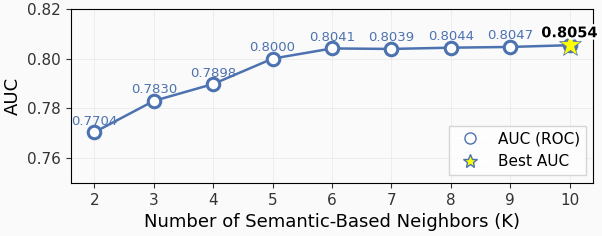}

\vspace{4pt}
\captionof{figure}{\textbf{Effect of semantic neighbors ($K$).}
Performance (AUC) improves as $K$ increases, and saturates beyond $K=6$,
indicating diminishing returns from additional neighbors.}
\label{fig:topk}

\vspace{6pt}

\centering
\footnotesize
\setlength{\tabcolsep}{10pt}
\renewcommand{\arraystretch}{1.3}

\resizebox{\textwidth}{!}{%
\begin{tabular}{c|cc|cc}
\toprule
\textbf{Top-$M$} & $\alpha_{\text{vis}}$ & $\alpha_{\text{text}}$ & \textbf{AUC} & \textbf{AP} \\
\midrule
10 & 0.5 & 0.5 & 0.7635 & 0.2907 \\
20 & 0.5 & 0.5 & 0.7994 & 0.3416 \\
25 & 0.5 & 0.5 & 0.8018 & 0.3636 \\
\rowcolor{green!10}
30 & 0.5 & 0.5 & \textbf{0.8054} & \textbf{0.3808} \\
50 & 0.5 & 0.5 & 0.8044 & 0.4012 \\
\midrule
$K=4$ & -- & -- & 0.7898 & 0.3347 \\
$K=6$ & -- & -- & 0.8041 & 0.3632 \\
\rowcolor{green!10}
$K=10$ & -- & -- & \textbf{0.8054} & \textbf{0.3808} \\
\bottomrule
\end{tabular}%
}

\vspace{2pt}
\captionsetup{font=scriptsize,skip=4pt}
\captionof{table}{
\textbf{Ablation of refinement parameters.}
Influence of $M$, temporal window $K$, and fusion weights on AUC and AP.
}
\label{tab:ablation_combined}
\end{minipage}
\vspace{5pt}
\\
\noindent Overall, these refinements harmonize semantic context, temporal consistency, and cross-modal alignment.
\textit{More ablations are provided in the Appendix \ref{sec:more_ab}}
\section{Concluding Remarks}
We presented Cog-VADU, a training-free framework for video anomaly detection and understanding that models anomalies through sequential cognitive reasoning over video segments. By leveraging a model-agnostic chain-of-thought prompting strategy with CoADTP for step-wise rationale propagation and a cross-modal re-ranking stage, Cog-VADU produces temporally coherent and interpretable anomaly scores without requiring fine-tuning. Experiments on UCF-Crime, XD-Violence, VANE, and HIVAU-70k demonstrate competitive zero-shot performance and strong generalization, highlighting the effectiveness of structured cognitive reasoning and causal understanding for real-world anomaly detection. In addition, we re-annotated two major datasets in a unified manner by incorporating explicit cause-and-effect relationships, which are crucial for analyzing anomalies in real-world scenarios.

\subsubsection*{Broader Impact Statement}
\label{sec:broder_impact}
This work demonstrates how the reasoning capabilities of foundation 
models can enable training-free video anomaly detection with 
interpretable outputs. By decomposing complex scenarios into 
structured steps, the approach improves both performance and trust in 
AI systems for video surveillance, such as public safety. It also 
shows strong potential for generalization in open-world settings. 
However, like all video surveillance technologies, deployment must 
carefully consider privacy, fairness, and responsible use. 
We discuss below the position Cog-VADU takes on four 
dimensions that we view as central to responsible deployment: 
privacy, false positives and over-surveillance, bias and fairness, 
and broader responsible-deployment guidelines.

\paragraph{Privacy.}
Cog-VADU is training-free and uses a frozen LVLM. Unlike supervised 
VAD systems that require large, labelled surveillance corpora,
which themselves raise consent, storage, and re-identification 
concerns, Cog-VADU introduces \emph{no} additional surveillance 
data collection or storage requirement for model adaptation. To 
support privacy-respecting deployment, we recommend that operators 
(i)~restrict inference to videos collected under documented consent 
or legal authorisation, (ii)~avoid persistent storage of frame-level 
rationales that could be used for re-identification, and 
(iii)~operate behind on-premise or institutionally-controlled 
inference rather than third-party APIs that may retain inputs.

\paragraph{False Positives and Over-Surveillance.}
Two properties of Cog-VADU are relevant to mitigating false alarms 
and over-surveillance. First, the framework produces a 
\emph{calibrated continuous score together with a textual rationale} 
for each segment, rather than a binary alert. This makes the system 
amenable to human-in-the-loop review, since operators can inspect 
the rationale before acting on a positive prediction. Second, our 
wrestling case study (Appendix~\ref{app:faithfulness}, 
Fig.~\ref{fig:case_wrestling}) shows that Cog-VADU correctly assigns 
a low score ($0.15$) to a visually high-motion-yet-benign scenario 
where motion-statistical methods 
(e.g., URDMU~\cite{zhou2023dualmemoryunitsuncertainty}) trigger 
near-saturated false alarms ($\approx 0.999$). Semantic grounding 
therefore directly reduces the false-alarm rate driven by motion 
statistics alone. We recommend that any deployment using Cog-VADU 
adopts a human-review protocol for positive predictions, and we 
explicitly disclaim its use as an autonomous alerting system.

\paragraph{Bias and Fairness.}
Cog-VADU inherits the biases of its underlying LVLM 
(VideoLLaMA3-7B~\cite{zhang2025videollama3frontiermultimodal}) and 
its anomaly database. While the training-free design avoids 
introducing additional dataset-specific bias from labelled 
surveillance corpora, it does not eliminate biases already present 
in the LVLM. The textual rationale, however, provides an 
\emph{auditability handle} not available in non-explainable 
baselines: biased decisions can be inspected and flagged at the 
explanation level, rather than emerging from an opaque score. We 
recommend that practitioners audit Cog-VADU's outputs against 
demographic and contextual subgroups before any real-world 
deployment, and document any subgroup disparities as part of the 
deployment record.

\paragraph{Responsible Deployment.}
We state explicitly that:
\begin{itemize}
    \item Cog-VADU is intended as a \emph{decision-support tool for 
    human operators}, not as an autonomous decision-making system.
    \item It should not be used for predictive policing, individual 
    identification, or any setting where its outputs directly 
    trigger consequential actions without human review.
    \item Known failure modes documented in this paper (e.g., the 
    categorical labelling failure illustrated in 
    Fig.~\ref{fig:failure_flood}) must be disclosed to end users so 
    that operators are aware of the framework's limitations.
    \item Use in high-stakes contexts (transportation safety, public 
    health emergencies) requires rigorous validation on the specific 
    deployment distribution before any reliance is placed on its 
    predictions.
\end{itemize}

In summary, Cog-VADU's training-free design and explainable outputs 
offer concrete advantages for privacy-respecting and auditable 
deployment, but these benefits depend on responsible deployment 
practices, human oversight, and recognition of the framework's 
documented failure modes.

\section{Code and Data Availability}

The code and re-annotated datasets associated with this work are publicly
available at \url{https://github.com/MohdUbaidwani/Cog-VADU}

\section*{Acknowledgements} 
We acknowledge the Barcelona Supercomputing Center (BSC) for providing computational resources on MareNostrum 5 through the EuroHPC Development Access project EHPC-DEV-2026D04-087. We also acknowledge the University of Surrey for its institutional and research support.


\bibliography{tmlr}

@misc{wu2024starbenchmarksituatedreasoning,
      title={STAR: A Benchmark for Situated Reasoning in Real-World Videos}, 
      author={Bo Wu and Shoubin Yu and Zhenfang Chen and Joshua B Tenenbaum and Chuang Gan},
      year={2024},
      eprint={2405.09711},
      archivePrefix={arXiv},
      primaryClass={cs.AI},
      url={https://arxiv.org/abs/2405.09711}, 
}

@inproceedings{zhang2025cause,
  title={Cause and Effect: Video Social Relationship Recognition from Causal Perspective},
  author={Zhang, Yuxuan and Wang, Bo and Du, Yu and Zhu, Yangfu and Wang, Haorui and Su, Guangyao and Zhou, Tao and Wu, Bin},
  booktitle={Proceedings of the 33rd ACM International Conference on Multimedia},
  pages={739--747},
  year={2025}
}

@InProceedings{Araghi_2025_CVPR,
    author    = {Araghi, Hesam and van Gemert, Jan and Tomen, Nergis},
    title     = {Making Every Event Count: Balancing Data Efficiency and Accuracy in Event Camera Subsampling},
    booktitle = {Proceedings of the IEEE/CVF Conference on Computer Vision and Pattern Recognition (CVPR) Workshops},
    month     = {June},
    year      = {2025},
    pages     = {5083-5093}
}

@misc{zhang2020bertscoreevaluatingtextgeneration,
      title={BERTScore: Evaluating Text Generation with BERT}, 
      author={Tianyi Zhang and Varsha Kishore and Felix Wu and Kilian Q. Weinberger and Yoav Artzi},
      year={2020},
      eprint={1904.09675},
      archivePrefix={arXiv},
      primaryClass={cs.CL},
      url={https://arxiv.org/abs/1904.09675}, 
}

@misc{zhang2024llavanext-video,
  title={LLaVA-NeXT: A Strong Zero-shot Video Understanding Model},
  url={https://llava-vl.github.io/blog/2024-04-30-llava-next-video/},
  author={Zhang, Yuanhan and Li, Bo and Liu, haotian and Lee, Yong jae and Gui, Liangke and Fu, Di and Feng, Jiashi and Liu, Ziwei and Li, Chunyuan},
  month={April},
  year={2024}
}

@misc{li2024videochatchatcentricvideounderstanding,
      title={VideoChat: Chat-Centric Video Understanding}, 
      author={KunChang Li and Yinan He and Yi Wang and Yizhuo Li and Wenhai Wang and Ping Luo and Yali Wang and Limin Wang and Yu Qiao},
      year={2024},
      eprint={2305.06355},
      archivePrefix={arXiv},
      primaryClass={cs.CV},
      url={https://arxiv.org/abs/2305.06355}, 
}

@misc{ren2024timechattimesensitivemultimodallarge,
      title={TimeChat: A Time-sensitive Multimodal Large Language Model for Long Video Understanding}, 
      author={Shuhuai Ren and Linli Yao and Shicheng Li and Xu Sun and Lu Hou},
      year={2024},
      eprint={2312.02051},
      archivePrefix={arXiv},
      primaryClass={cs.CV},
      url={https://arxiv.org/abs/2312.02051}, 
}

@misc{zaheer2021clawsclusteringassistedweakly,
      title={CLAWS: Clustering Assisted Weakly Supervised Learning with Normalcy Suppression for Anomalous Event Detection}, 
      author={Muhammad Zaigham Zaheer and Arif Mahmood and Marcella Astrid and Seung-Ik Lee},
      year={2021},
      eprint={2011.12077},
      archivePrefix={arXiv},
      primaryClass={cs.CV},
      url={https://arxiv.org/abs/2011.12077}, 
}

@article{feng2025video,
  title={Video-R1: Reinforcing Video Reasoning in MLLMs},
  author={Feng, Kaituo and Gong, Kaixiong and Li, Bohao and Guo, Zonghao and Wang, Yibing and Peng, Tianshuo and Wang, Benyou and Yue, Xiangyu},
  journal={arXiv preprint arXiv:2503.21776},
  year={2025}
}

@misc{zhu2025vaur1advancingvideoanomaly,
      title={VAU-R1: Advancing Video Anomaly Understanding via Reinforcement Fine-Tuning}, 
      author={Liyun Zhu and Qixiang Chen and Xi Shen and Xiaodong Cun},
      year={2025},
      eprint={2505.23504},
      archivePrefix={arXiv},
      primaryClass={cs.CV},
      url={https://arxiv.org/abs/2505.23504}, 
}

@inproceedings{chen2024internvl,
  title={Internvl: Scaling up vision foundation models and aligning for generic visual-linguistic tasks},
  author={Chen, Zhe and Wu, Jiannan and Wang, Wenhai and Su, Weijie and Chen, Guo and Xing, Sen and Zhong, Muyan and Zhang, Qinglong and Zhu, Xizhou and Lu, Lewei and others},
  booktitle={Proceedings of the IEEE/CVF Conference on Computer Vision and Pattern Recognition},
  pages={24185--24198},
  year={2024}
}

@misc{wang2024qwen2vlenhancingvisionlanguagemodels,
      title={Qwen2-VL: Enhancing Vision-Language Model's Perception of the World at Any Resolution}, 
      author={Peng Wang and Shuai Bai and Sinan Tan and Shijie Wang and Zhihao Fan and Jinze Bai and Keqin Chen and Xuejing Liu and Jialin Wang and Wenbin Ge and Yang Fan and Kai Dang and Mengfei Du and Xuancheng Ren and Rui Men and Dayiheng Liu and Chang Zhou and Jingren Zhou and Junyang Lin},
      year={2024},
      eprint={2409.12191},
      archivePrefix={arXiv},
      primaryClass={cs.CV},
      url={https://arxiv.org/abs/2409.12191}, 
}

@misc{maaz2024videochatgptdetailedvideounderstanding,
      title={Video-ChatGPT: Towards Detailed Video Understanding via Large Vision and Language Models}, 
      author={Muhammad Maaz and Hanoona Rasheed and Salman Khan and Fahad Shahbaz Khan},
      year={2024},
      eprint={2306.05424},
      archivePrefix={arXiv},
      primaryClass={cs.CV},
      url={https://arxiv.org/abs/2306.05424}, 
}

@misc{bharadwaj2025vanebenchvideoanomalyevaluation,
      title={VANE-Bench: Video Anomaly Evaluation Benchmark for Conversational LMMs}, 
      author={Rohit Bharadwaj and Hanan Gani and Muzammal Naseer and Fahad Shahbaz Khan and Salman Khan},
      year={2025},
      eprint={2406.10326},
      archivePrefix={arXiv},
      primaryClass={cs.CV},
      url={https://arxiv.org/abs/2406.10326}, 
}

@misc{wan2021weaklysupervisedvideoanomaly,
      title={Weakly Supervised Video Anomaly Detection via Center-guided Discriminative Learning}, 
      author={Boyang Wan and Yuming Fang and Xue Xia and Jiajie Mei},
      year={2021},
      eprint={2104.07268},
      archivePrefix={arXiv},
      primaryClass={cs.CV},
      url={https://arxiv.org/abs/2104.07268}, 
}

@misc{xiao2021nextqanextphasequestionansweringexplaining,
      title={NExT-QA:Next Phase of Question-Answering to Explaining Temporal Actions}, 
      author={Junbin Xiao and Xindi Shang and Angela Yao and Tat-Seng Chua},
      year={2021},
      eprint={2105.08276},
      archivePrefix={arXiv},
      primaryClass={cs.CV},
      url={https://arxiv.org/abs/2105.08276}, 
}

@misc{yang2023videoeventrestorationbased,
      title={Video Event Restoration Based on Keyframes for Video Anomaly Detection}, 
      author={Zhiwei Yang and Jing Liu and Zhaoyang Wu and Peng Wu and Xiaotao Liu},
      year={2023},
      eprint={2304.05112},
      archivePrefix={arXiv},
      primaryClass={cs.CV},
      url={https://arxiv.org/abs/2304.05112}, 
}

@misc{huang2025vadr1videoanomalyreasoning,
      title={Vad-R1: Towards Video Anomaly Reasoning via Perception-to-Cognition Chain-of-Thought}, 
      author={Chao Huang and Benfeng Wang and Jie Wen and Chengliang Liu and Wei Wang and Li Shen and Xiaochun Cao},
      year={2025},
      eprint={2505.19877},
      archivePrefix={arXiv},
      primaryClass={cs.CV},
      url={https://arxiv.org/abs/2505.19877}, 
}

@misc{gong2019memorizingnormalitydetectanomaly,
      title={Memorizing Normality to Detect Anomaly: Memory-augmented Deep Autoencoder for Unsupervised Anomaly Detection}, 
      author={Dong Gong and Lingqiao Liu and Vuong Le and Budhaditya Saha and Moussa Reda Mansour and Svetha Venkatesh and Anton van den Hengel},
      year={2019},
      eprint={1904.02639},
      archivePrefix={arXiv},
      primaryClass={cs.CV},
      url={https://arxiv.org/abs/1904.02639}, 
}

@misc{thakare2022dyannetscenedynamicityguided,
      title={DyAnNet: A Scene Dynamicity Guided Self-Trained Video Anomaly Detection Network}, 
      author={Kamalakar Thakare and Yash Raghuwanshi and Debi Prosad Dogra and Heeseung Choi and Ig-Jae Kim},
      year={2022},
      eprint={2211.00882},
      archivePrefix={arXiv},
      primaryClass={cs.CV},
      url={https://arxiv.org/abs/2211.00882}, 
}

@misc{fei2024videoofthoughtstepbystepvideoreasoning,
      title={Video-of-Thought: Step-by-Step Video Reasoning from Perception to Cognition}, 
      author={Hao Fei and Shengqiong Wu and Wei Ji and Hanwang Zhang and Meishan Zhang and Mong-Li Lee and Wynne Hsu},
      year={2024},
      eprint={2501.03230},
      archivePrefix={arXiv},
      primaryClass={cs.AI},
      url={https://arxiv.org/abs/2501.03230}, 
}

@misc{zhang2022automaticchainthoughtprompting,
      title={Automatic Chain of Thought Prompting in Large Language Models}, 
      author={Zhuosheng Zhang and Aston Zhang and Mu Li and Alex Smola},
      year={2022},
      eprint={2210.03493},
      archivePrefix={arXiv},
      primaryClass={cs.CL},
      url={https://arxiv.org/abs/2210.03493}, 
}

@misc{yang2025monitorexploitinglargelanguage,
      title={MoniTor: Exploiting Large Language Models with Instruction for Online Video Anomaly Detection}, 
      author={Shengtian Yang and Yue Feng and Yingshi Liu and Jingrou Zhang and Jie Qin},
      year={2025},
      eprint={2510.21449},
      archivePrefix={arXiv},
      primaryClass={cs.CV},
      url={https://arxiv.org/abs/2510.21449}, 
}

@misc{ma2025sherlockmultiscenevideoabnormal,
      title={Sherlock: Towards Multi-scene Video Abnormal Event Extraction and Localization via a Global-local Spatial-sensitive LLM}, 
      author={Junxiao Ma and Jingjing Wang and Jiamin Luo and Peiying Yu and Guodong Zhou},
      year={2025},
      eprint={2502.18863},
      archivePrefix={arXiv},
      primaryClass={cs.CV},
      url={https://arxiv.org/abs/2502.18863}, 
}

@ARTICLE{9591368,
  author={Huang, Chao and Wu, Zhihao and Wen, Jie and Xu, Yong and Jiang, Qiuping and Wang, Yaowei},
  journal={IEEE Transactions on Industrial Informatics}, 
  title={Abnormal Event Detection Using Deep Contrastive Learning for Intelligent Video Surveillance System}, 
  year={2022},
  volume={18},
  number={8},
  pages={5171-5179},
  doi={10.1109/TII.2021.3122801}}

@misc{liu2018futureframepredictionanomaly,
      title={Future Frame Prediction for Anomaly Detection -- A New Baseline}, 
      author={Wen Liu and Weixin Luo and Dongze Lian and Shenghua Gao},
      year={2018},
      eprint={1712.09867},
      archivePrefix={arXiv},
      primaryClass={cs.CV},
      url={https://arxiv.org/abs/1712.09867}, 
}

@String(CVPR= {IEEE Conf. Comput. Vis. Pattern Recog.})

@String(ICIP = {IEEE Int. Conf. Image Process.})

@String(CVPRW= {IEEE Conf. Comput. Vis. Pattern Recog. Worksh.})

@String(CVPR  = {CVPR})

@String(ICIP  = {ICIP})

@String(CVPRW= {CVPRW})

@misc{li2023blip2bootstrappinglanguageimagepretraining,
      title={BLIP-2: Bootstrapping Language-Image Pre-training with Frozen Image Encoders and Large Language Models}, 
      author={Junnan Li and Dongxu Li and Silvio Savarese and Steven Hoi},
      year={2023},
      eprint={2301.12597},
      archivePrefix={arXiv},
      primaryClass={cs.CV},
      url={https://arxiv.org/abs/2301.12597}, 
}

@inproceedings{wang2010crowd,
author = {Wang, Shu and Miao, Zhenjiang},
year = {2010},
month = {11},
pages = {1220 - 1223},
title = {Anomaly detection in crowd scene},
doi = {10.1109/ICOSP.2010.5655356}
}

@INPROCEEDINGS{10208872,
  author={Chen, Weiling and Ma, Keng Teck and Jian Yew, Zi and Hur, Minhoe and Khoo, David Aik-Aun},
  booktitle={2023 IEEE/CVF Conference on Computer Vision and Pattern Recognition Workshops (CVPRW)}, 
  title={TEVAD: Improved video anomaly detection with captions}, 
  year={2023},
  volume={},
  number={},
  pages={5549-5559},
  doi={10.1109/CVPRW59228.2023.00587}}

@misc{tur2023exploringdiffusionmodelsunsupervised,
      title={Exploring Diffusion Models for Unsupervised Video Anomaly Detection}, 
      author={Anil Osman Tur and Nicola Dall'Asen and Cigdem Beyan and Elisa Ricci},
      year={2023},
      eprint={2304.05841},
      archivePrefix={arXiv},
      primaryClass={cs.CV},
      url={https://arxiv.org/abs/2304.05841}, 
}

@misc{chen2024internvlscalingvisionfoundation,
      title={InternVL: Scaling up Vision Foundation Models and Aligning for Generic Visual-Linguistic Tasks}, 
      author={Zhe Chen and Jiannan Wu and Wenhai Wang and Weijie Su and Guo Chen and Sen Xing and Muyan Zhong and Qinglong Zhang and Xizhou Zhu and Lewei Lu and Bin Li and Ping Luo and Tong Lu and Yu Qiao and Jifeng Dai},
      year={2024},
      eprint={2312.14238},
      archivePrefix={arXiv},
      primaryClass={cs.CV},
      url={https://arxiv.org/abs/2312.14238}, 
}

@misc{li2023llamavidimageworth2,
      title={LLaMA-VID: An Image is Worth 2 Tokens in Large Language Models}, 
      author={Yanwei Li and Chengyao Wang and Jiaya Jia},
      year={2023},
      eprint={2311.17043},
      archivePrefix={arXiv},
      primaryClass={cs.CV},
      url={https://arxiv.org/abs/2311.17043}, 
}

@misc{song2024moviechatdensetokensparse,
      title={MovieChat: From Dense Token to Sparse Memory for Long Video Understanding}, 
      author={Enxin Song and Wenhao Chai and Guanhong Wang and Yucheng Zhang and Haoyang Zhou and Feiyang Wu and Haozhe Chi and Xun Guo and Tian Ye and Yanting Zhang and Yan Lu and Jenq-Neng Hwang and Gaoang Wang},
      year={2024},
      eprint={2307.16449},
      archivePrefix={arXiv},
      primaryClass={cs.CV},
      url={https://arxiv.org/abs/2307.16449}, 
}

@misc{tang2024hawklearningunderstandopenworld,
      title={Hawk: Learning to Understand Open-World Video Anomalies}, 
      author={Jiaqi Tang and Hao Lu and Ruizheng Wu and Xiaogang Xu and Ke Ma and Cheng Fang and Bin Guo and Jiangbo Lu and Qifeng Chen and Ying-Cong Chen},
      year={2024},
      eprint={2405.16886},
      archivePrefix={arXiv},
      primaryClass={cs.CV},
      url={https://arxiv.org/abs/2405.16886}, 
}

@misc{ye2025veraexplainablevideoanomaly,
      title={VERA: Explainable Video Anomaly Detection via Verbalized Learning of Vision-Language Models}, 
      author={Muchao Ye and Weiyang Liu and Pan He},
      year={2025},
      eprint={2412.01095},
      archivePrefix={arXiv},
      primaryClass={cs.AI},
      url={https://arxiv.org/abs/2412.01095}, 
}

@inproceedings{joo2023clip,
  title={Clip-tsa: Clip-assisted temporal self-attention for weakly-supervised video anomaly detection},
  author={Joo, Hyekang Kevin and Vo, Khoa and Yamazaki, Kashu and Le, Ngan},
  booktitle={2023 IEEE International Conference on Image Processing (ICIP)},
  pages={3230--3234},
  year={2023},
  organization={IEEE}
}

@inproceedings{li2022scale,
  title={Scale-aware spatio-temporal relation learning for video anomaly detection},
  author={Li, Guoqiu and Cai, Guanxiong and Zeng, Xingyu and Zhao, Rui},
  booktitle={European Conference on Computer Vision},
  pages={333--350},
  year={2022},
  organization={Springer}
}

@misc{yuan2023surveillancevideoandlanguageunderstandingnew,
      title={Towards Surveillance Video-and-Language Understanding: New Dataset, Baselines, and Challenges}, 
      author={Tongtong Yuan and Xuange Zhang and Kun Liu and Bo Liu and Chen Chen and Jian Jin and Zhenzhen Jiao},
      year={2023},
      eprint={2309.13925},
      archivePrefix={arXiv},
      primaryClass={cs.CV},
      url={https://arxiv.org/abs/2309.13925}, 
}

@misc{zhang2025holmesvaulongtermvideoanomaly,
      title={Holmes-VAU: Towards Long-term Video Anomaly Understanding at Any Granularity}, 
      author={Huaxin Zhang and Xiaohao Xu and Xiang Wang and Jialong Zuo and Xiaonan Huang and Changxin Gao and Shanjun Zhang and Li Yu and Nong Sang},
      year={2025},
      eprint={2412.06171},
      archivePrefix={arXiv},
      primaryClass={cs.CV},
      url={https://arxiv.org/abs/2412.06171}, 
}

@misc{yang2024textpromptnormalityguidance,
      title={Text Prompt with Normality Guidance for Weakly Supervised Video Anomaly Detection}, 
      author={Zhiwei Yang and Jing Liu and Peng Wu},
      year={2024},
      eprint={2404.08531},
      archivePrefix={arXiv},
      primaryClass={cs.CV},
      url={https://arxiv.org/abs/2404.08531}, 
}

@misc{chen2022mgfnmagnitudecontrastiveglanceandfocusnetwork,
      title={MGFN: Magnitude-Contrastive Glance-and-Focus Network for Weakly-Supervised Video Anomaly Detection}, 
      author={Yingxian Chen and Zhengzhe Liu and Baoheng Zhang and Wilton Fok and Xiaojuan Qi and Yik-Chung Wu},
      year={2022},
      eprint={2211.15098},
      archivePrefix={arXiv},
      primaryClass={cs.CV},
      url={https://arxiv.org/abs/2211.15098}, 
}

@misc{pu2024learningpromptenhancedcontextfeatures,
      title={Learning Prompt-Enhanced Context Features for Weakly-Supervised Video Anomaly Detection}, 
      author={Yujiang Pu and Xiaoyu Wu and Lulu Yang and Shengjin Wang},
      year={2024},
      eprint={2306.14451},
      archivePrefix={arXiv},
      primaryClass={cs.CV},
      url={https://arxiv.org/abs/2306.14451}, 
}

@misc{wu2023vadclipadaptingvisionlanguagemodels,
      title={VadCLIP: Adapting Vision-Language Models for Weakly Supervised Video Anomaly Detection}, 
      author={Peng Wu and Xuerong Zhou and Guansong Pang and Lingru Zhou and Qingsen Yan and Peng Wang and Yanning Zhang},
      year={2023},
      eprint={2308.11681},
      archivePrefix={arXiv},
      primaryClass={cs.CV},
      url={https://arxiv.org/abs/2308.11681}, 
}

@article{10.1016/j.patcog.2023.109567,
    author = {Thakare, Kamalakar Vijay and Dogra, Debi Prosad and Choi, Heeseung and Kim, Haksub and Kim, Ig-Jae},
    title = {RareAnom: A Benchmark Video Dataset for Rare Type Anomalies},
    year = {2023},
    issue_date = {Aug 2023},
    publisher = {Elsevier Science Inc.},
    address = {USA},
    volume = {140},
    number = {C},
    issn = {0031-3203},
    url = {https://doi.org/10.1016/j.patcog.2023.109567},
    doi = {10.1016/j.patcog.2023.109567},
    journal = {Pattern Recogn.},
    month = aug,
    numpages = {12}
    }

@misc{zhang2024glancevadexploringglancesupervision,
      title={GlanceVAD: Exploring Glance Supervision for Label-efficient Video Anomaly Detection}, 
      author={Huaxin Zhang and Xiang Wang and Xiaohao Xu and Xiaonan Huang and Chuchu Han and Yuehuan Wang and Changxin Gao and Shanjun Zhang and Nong Sang},
      year={2024},
      eprint={2403.06154},
      archivePrefix={arXiv},
      primaryClass={cs.CV},
      url={https://arxiv.org/abs/2403.06154}, 
}

@misc{wang2019godsgeneralizedoneclassdiscriminative,
      title={GODS: Generalized One-class Discriminative Subspaces for Anomaly Detection}, 
      author={Jue Wang and Anoop Cherian},
      year={2019},
      eprint={1908.05884},
      archivePrefix={arXiv},
      primaryClass={cs.CV},
      url={https://arxiv.org/abs/1908.05884}, 
}

@misc{liu2024improvedbaselinesvisualinstruction,
      title={Improved Baselines with Visual Instruction Tuning}, 
      author={Haotian Liu and Chunyuan Li and Yuheng Li and Yong Jae Lee},
      year={2024},
      eprint={2310.03744},
      archivePrefix={arXiv},
      primaryClass={cs.CV},
      url={https://arxiv.org/abs/2310.03744}, 
}

@misc{wei2023chainofthoughtpromptingelicitsreasoning,
      title={Chain-of-Thought Prompting Elicits Reasoning in Large Language Models}, 
      author={Jason Wei and Xuezhi Wang and Dale Schuurmans and Maarten Bosma and Brian Ichter and Fei Xia and Ed Chi and Quoc Le and Denny Zhou},
      year={2023},
      eprint={2201.11903},
      archivePrefix={arXiv},
      primaryClass={cs.CL},
      url={https://arxiv.org/abs/2201.11903}, 
}

@misc{wu2020looklistenlearningmultimodal,
      title={Not only Look, but also Listen: Learning Multimodal Violence Detection under Weak Supervision}, 
      author={Peng Wu and Jing Liu and Yujia Shi and Yujia Sun and Fangtao Shao and Zhaoyang Wu and Zhiwei Yang},
      year={2020},
      eprint={2007.04687},
      archivePrefix={arXiv},
      primaryClass={cs.CV},
      url={https://arxiv.org/abs/2007.04687}, 
}

@misc{lin2024videollavalearningunitedvisual,
      title={Video-LLaVA: Learning United Visual Representation by Alignment Before Projection}, 
      author={Bin Lin and Yang Ye and Bin Zhu and Jiaxi Cui and Munan Ning and Peng Jin and Li Yuan},
      year={2024},
      eprint={2311.10122},
      archivePrefix={arXiv},
      primaryClass={cs.CV},
      url={https://arxiv.org/abs/2311.10122}, 
}

@misc{zhang2024holmesvadunbiasedexplainablevideo,
      title={Holmes-VAD: Towards Unbiased and Explainable Video Anomaly Detection via Multi-modal LLM}, 
      author={Huaxin Zhang and Xiaohao Xu and Xiang Wang and Jialong Zuo and Chuchu Han and Xiaonan Huang and Changxin Gao and Yuehuan Wang and Nong Sang},
      year={2024},
      eprint={2406.12235},
      archivePrefix={arXiv},
      primaryClass={cs.CV},
      url={https://arxiv.org/abs/2406.12235}, 
}

@misc{arnab2025temporalchainthoughtlongvideo,
      title={Temporal Chain of Thought: Long-Video Understanding by Thinking in Frames}, 
      author={Anurag Arnab and Ahmet Iscen and Mathilde Caron and Alireza Fathi and Cordelia Schmid},
      year={2025},
      eprint={2507.02001},
      archivePrefix={arXiv},
      primaryClass={cs.LG},
      url={https://arxiv.org/abs/2507.02001}, 
}

@misc{zhang2025videocotcomprehensivedatasetspatiotemporal,
      title={Video-CoT: A Comprehensive Dataset for Spatiotemporal Understanding of Videos Based on Chain-of-Thought}, 
      author={Shuyi Zhang and Xiaoshuai Hao and Yingbo Tang and Lingfeng Zhang and Pengwei Wang and Zhongyuan Wang and Hongxuan Ma and Shanghang Zhang},
      year={2025},
      eprint={2506.08817},
      archivePrefix={arXiv},
      primaryClass={cs.CV},
      url={https://arxiv.org/abs/2506.08817}, 
}

@misc{zhang2024improvevisionlanguagemodel,
      title={Improve Vision Language Model Chain-of-thought Reasoning}, 
      author={Ruohong Zhang and Bowen Zhang and Yanghao Li and Haotian Zhang and Zhiqing Sun and Zhe Gan and Yinfei Yang and Ruoming Pang and Yiming Yang},
      year={2024},
      eprint={2410.16198},
      archivePrefix={arXiv},
      primaryClass={cs.AI},
      url={https://arxiv.org/abs/2410.16198}, 
}

@misc{tian2021weaklysupervisedvideoanomalydetection,
      title={Weakly-supervised Video Anomaly Detection with Robust Temporal Feature Magnitude Learning}, 
      author={Yu Tian and Guansong Pang and Yuanhong Chen and Rajvinder Singh and Johan W. Verjans and Gustavo Carneiro},
      year={2021},
      eprint={2101.10030},
      archivePrefix={arXiv},
      primaryClass={cs.CV},
      url={https://arxiv.org/abs/2101.10030}, 
}

@misc{zhou2023dualmemoryunitsuncertainty,
      title={Dual Memory Units with Uncertainty Regulation for Weakly Supervised Video Anomaly Detection}, 
      author={Hang Zhou and Junqing Yu and Wei Yang},
      year={2023},
      eprint={2302.05160},
      archivePrefix={arXiv},
      primaryClass={cs.CV},
      url={https://arxiv.org/abs/2302.05160}, 
}

@misc{zhang2025videollama3frontiermultimodal,
      title={VideoLLaMA 3: Frontier Multimodal Foundation Models for Image and Video Understanding}, 
      author={Boqiang Zhang and Kehan Li and Zesen Cheng and Zhiqiang Hu and Yuqian Yuan and Guanzheng Chen and Sicong Leng and Yuming Jiang and Hang Zhang and Xin Li and Peng Jin and Wenqi Zhang and Fan Wang and Lidong Bing and Deli Zhao},
      year={2025},
      eprint={2501.13106},
      archivePrefix={arXiv},
      primaryClass={cs.CV},
      url={https://arxiv.org/abs/2501.13106}, 
}

@misc{nadeem2025narrativebridgeenhancingvideocaptioning,
      title={NarrativeBridge: Enhancing Video Captioning with Causal-Temporal Narrative}, 
      author={Asmar Nadeem and Faegheh Sardari and Robert Dawes and Syed Sameed Husain and Adrian Hilton and Armin Mustafa},
      year={2025},
      eprint={2406.06499},
      archivePrefix={arXiv},
      primaryClass={cs.CV},
      url={https://arxiv.org/abs/2406.06499}, 
}

@misc{girdhar2023imagebindembeddingspacebind,
      title={ImageBind: One Embedding Space To Bind Them All}, 
      author={Rohit Girdhar and Alaaeldin El-Nouby and Zhuang Liu and Mannat Singh and Kalyan Vasudev Alwala and Armand Joulin and Ishan Misra},
      year={2023},
      eprint={2305.05665},
      archivePrefix={arXiv},
      primaryClass={cs.CV},
      url={https://arxiv.org/abs/2305.05665}, 
}

@misc{zhang2023videollamainstructiontunedaudiovisuallanguage,
      title={Video-LLaMA: An Instruction-tuned Audio-Visual Language Model for Video Understanding}, 
      author={Hang Zhang and Xin Li and Lidong Bing},
      year={2023},
      eprint={2306.02858},
      archivePrefix={arXiv},
      primaryClass={cs.CL},
      url={https://arxiv.org/abs/2306.02858}, 
}

@misc{hasan2016learningtemporalregularityvideo,
      title={Learning Temporal Regularity in Video Sequences}, 
      author={Mahmudul Hasan and Jonghyun Choi and Jan Neumann and Amit K. Roy-Chowdhury and Larry S. Davis},
      year={2016},
      eprint={1604.04574},
      archivePrefix={arXiv},
      primaryClass={cs.CV},
      url={https://arxiv.org/abs/1604.04574}, 
}

@misc{zanella2024harnessinglargelanguagemodels,
      title={Harnessing Large Language Models for Training-free Video Anomaly Detection}, 
      author={Luca Zanella and Willi Menapace and Massimiliano Mancini and Yiming Wang and Elisa Ricci},
      year={2024},
      eprint={2404.01014},
      archivePrefix={arXiv},
      primaryClass={cs.CV},
      url={https://arxiv.org/abs/2404.01014}, 
}

@misc{sultani2019realworldanomalydetectionsurveillance,
      title={Real-world Anomaly Detection in Surveillance Videos}, 
      author={Waqas Sultani and Chen Chen and Mubarak Shah},
      year={2019},
      eprint={1801.04264},
      archivePrefix={arXiv},
      primaryClass={cs.CV},
      url={https://arxiv.org/abs/1801.04264}, 
}

@misc{lv2024videoanomalydetectionexplanation,
      title={Video Anomaly Detection and Explanation via Large Language Models}, 
      author={Hui Lv and Qianru Sun},
      year={2024},
      eprint={2401.05702},
      archivePrefix={arXiv},
      primaryClass={cs.CV},
      url={https://arxiv.org/abs/2401.05702}, 
}

@misc{zhou2023batchnormbasedweaklysupervisedvideo,
      title={BatchNorm-based Weakly Supervised Video Anomaly Detection}, 
      author={Yixuan Zhou and Yi Qu and Xing Xu and Fumin Shen and Jingkuan Song and Hengtao Shen},
      year={2023},
      eprint={2311.15367},
      archivePrefix={arXiv},
      primaryClass={cs.CV},
      url={https://arxiv.org/abs/2311.15367}, 
}

@misc{gao2025steadspatiotemporalefficientanomaly,
      title={STEAD: Spatio-Temporal Efficient Anomaly Detection for Time and Compute Sensitive Applications}, 
      author={Andrew Gao and Jun Liu},
      year={2025},
      eprint={2503.07942},
      archivePrefix={arXiv},
      primaryClass={cs.CV},
      url={https://arxiv.org/abs/2503.07942}, 
}

@misc{ilse2018attentionbaseddeepmultipleinstance,
      title={Attention-based Deep Multiple Instance Learning}, 
      author={Maximilian Ilse and Jakub M. Tomczak and Max Welling},
      year={2018},
      eprint={1802.04712},
      archivePrefix={arXiv},
      primaryClass={cs.LG},
      url={https://arxiv.org/abs/1802.04712}, 
}

@ARTICLE{5562666,
  author={Saligrama, Venkatesh and Konrad, Janusz and Jodoin, Pierre-marc},
  journal={IEEE Signal Processing Magazine}, 
  title={Video Anomaly Identification}, 
  year={2010},
  volume={27},
  number={5},
  pages={18-33},
  doi={10.1109/MSP.2010.937393}}

@inproceedings{10.1145/3343031.3350998,
author = {Liu, Kun and Ma, Huadong},
title = {Exploring Background-bias for Anomaly Detection in Surveillance Videos},
year = {2019},
isbn = {9781450368896},
publisher = {Association for Computing Machinery},
address = {New York, NY, USA},
url = {https://doi.org/10.1145/3343031.3350998},
doi = {10.1145/3343031.3350998},
booktitle = {Proceedings of the 27th ACM International Conference on Multimedia},
pages = {1490–1499},
numpages = {10},
location = {Nice, France},
series = {MM '19}
}

@misc{lv2023unbiasedmultipleinstancelearning,
      title={Unbiased Multiple Instance Learning for Weakly Supervised Video Anomaly Detection}, 
      author={Hui Lv and Zhongqi Yue and Qianru Sun and Bin Luo and Zhen Cui and Hanwang Zhang},
      year={2023},
      eprint={2303.12369},
      archivePrefix={arXiv},
      primaryClass={cs.CV},
      url={https://arxiv.org/abs/2303.12369}, 
}
\bibliographystyle{tmlr}

\appendix
\newpage
\section{Appendix}

This appendix provides supplementary information for the main paper. 
Firstly, we provide detailed information about the proposed unified re-annotated dataset, including the limitations of current annotations, the construction process, statistical analysis, and examples. 
Then, we present additional experimental details, including prompt design and model ablations. 
Finally, we provide extended results, visualizations, and discuss limitations and future directions.


\section{A Unified Annotation Protocol}
\label{sec:ANOT_protocol}

\subsection{Motivation and Benchmark Limitations}
\label{subsec:moti}
Recent advances in video understanding~\cite{zhang2025cause,nadeem2025narrativebridgeenhancingvideocaptioning,Araghi_2025_CVPR} highlight the importance of modeling causal and temporal relationships, rather than relying solely on instantaneous visual cues.
Benchmarks such as NExT-QA~\cite{xiao2021nextqanextphasequestionansweringexplaining} and STAR~\cite{wu2024starbenchmarksituatedreasoning} explicitly evaluate cause-effect reasoning, emphasizing that meaningful video interpretation often depends on contextual and temporal dependencies.
This perspective suggests that anomaly detection should extend beyond visually salient triggers (e.g., explosions) to include semantically meaningful states that follow such events.

Motivated by this, we propose a unified re-annotation protocol to better capture context-driven anomalies that are often overlooked in existing VAD benchmarks.
Through analysis of XD-Violence~\cite{wu2020looklistenlearningmultimodal} and UCF-Crime~\cite{sultani2019realworldanomalydetectionsurveillance}, we identify key annotation limitations.
Current annotations primarily focus on \emph{cause-based} anomalies, while frequently omitting \emph{effect-based} or context-dependent states, such as post-incident outcomes or prolonged abnormal situations.

These omissions introduce a structural bias.
Training-based methods tend to ignore subtle or contextual anomalies to avoid false positives, whereas reasoning-based models detect them but are penalized under incomplete ground truth.
This discrepancy contributes to reduced evaluation performance, particularly in metrics such as AP.

\begin{figure}[htbp]
\centering
\captionsetup{font=normalsize, skip=6pt}
\includegraphics[width=1\textwidth]{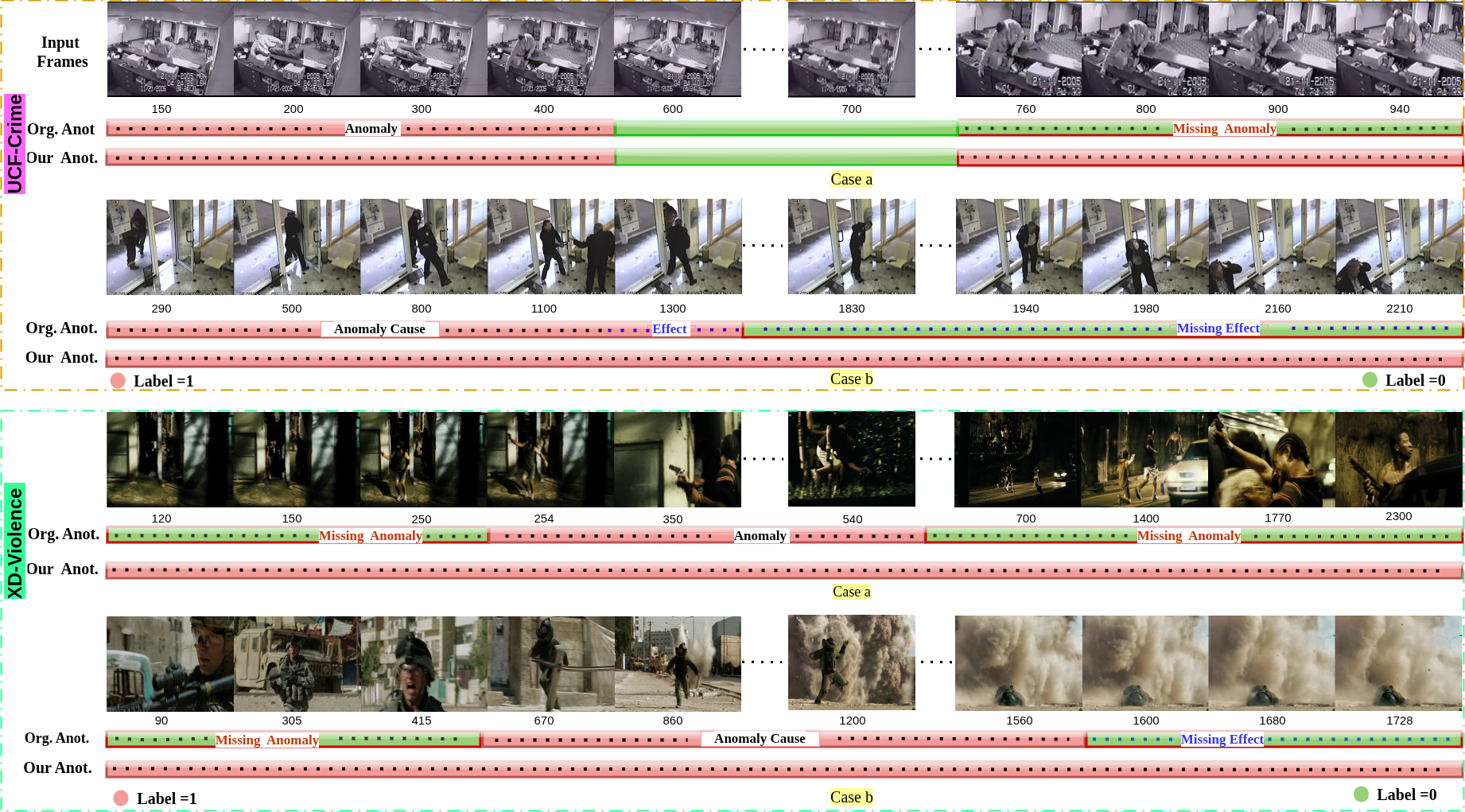}
\caption{\textbf{UCF~\cite{sultani2019realworldanomalydetectionsurveillance} and XD-Violence~\cite{wu2020looklistenlearningmultimodal} Re-Annotation Examples.}  
Case (a) shows missing anomaly regions in the original annotations but included in our unified annotations (\textbf{orig.Anot: Original Annotations, Our. Anot: Ours Re-Annotations}).  
Case (b) highlights cause–and–effect anomaly segments previously ignored but included in our unified annotations.}
\label{fig:my_anott}
\end{figure}

We further observe incomplete and inconsistent annotations, where repeated anomalies are labeled only once, and temporally dependent segments are partially or entirely missed.
As illustrated in Fig.~\ref{fig:my_anott}, repeated anomalous events are often annotated only at their first occurrence, while subsequent identical events remain unlabeled. Similarly, in Case~b of Fig.~\ref{fig:my_anott}, the causal event triggering the anomaly is annotated, whereas the subsequent effect segments, despite representing the same anomalous outcome, are omitted.

To address this, our unified protocol extends annotations to include these effect-based and temporally consistent anomaly segments.
This results in more complete anomaly coverage and improved temporal consistency.
Using this protocol, we construct refined annotations for both UCF-Crime~\cite{sultani2019realworldanomalydetectionsurveillance} and XD-Violence~\cite{wu2020looklistenlearningmultimodal}.
\subsection{Systematic Annotation Errors}
\label{subsec:syst}
Our frame-level analysis reveals three recurring issues:

\textbf{(1) Missed Effect Regions.}  
Only the initiating cause is labeled, while downstream consequences (e.g., injured victims, smoke-filled environments, evacuation scenes) remain unlabeled.

\textbf{(2) Repetition Bias.}  
Repeated anomalous actions are frequently annotated only once, ignoring later identical occurrences.

\textbf{(3) Inconsistent Cause-Effect Handling.}  
Videos containing causal chains (e.g., throwing an object $\rightarrow$ explosion $\rightarrow$ evacuation) are inconsistently labeled, sometimes marking only the cause, sometimes only the effect, and sometimes neither comprehensively.

\noindent These inconsistencies are substantially more prevalent in XD-Violence~\cite{wu2020looklistenlearningmultimodal}, explaining the larger performance gain we observe after correction.

\subsection{Unified Annotation Principle}
\label{subsec:principle}
\begin{figure*}
    \centering
    \captionsetup{font=normalsize, skip=6pt}\includegraphics[width=1.0\linewidth]{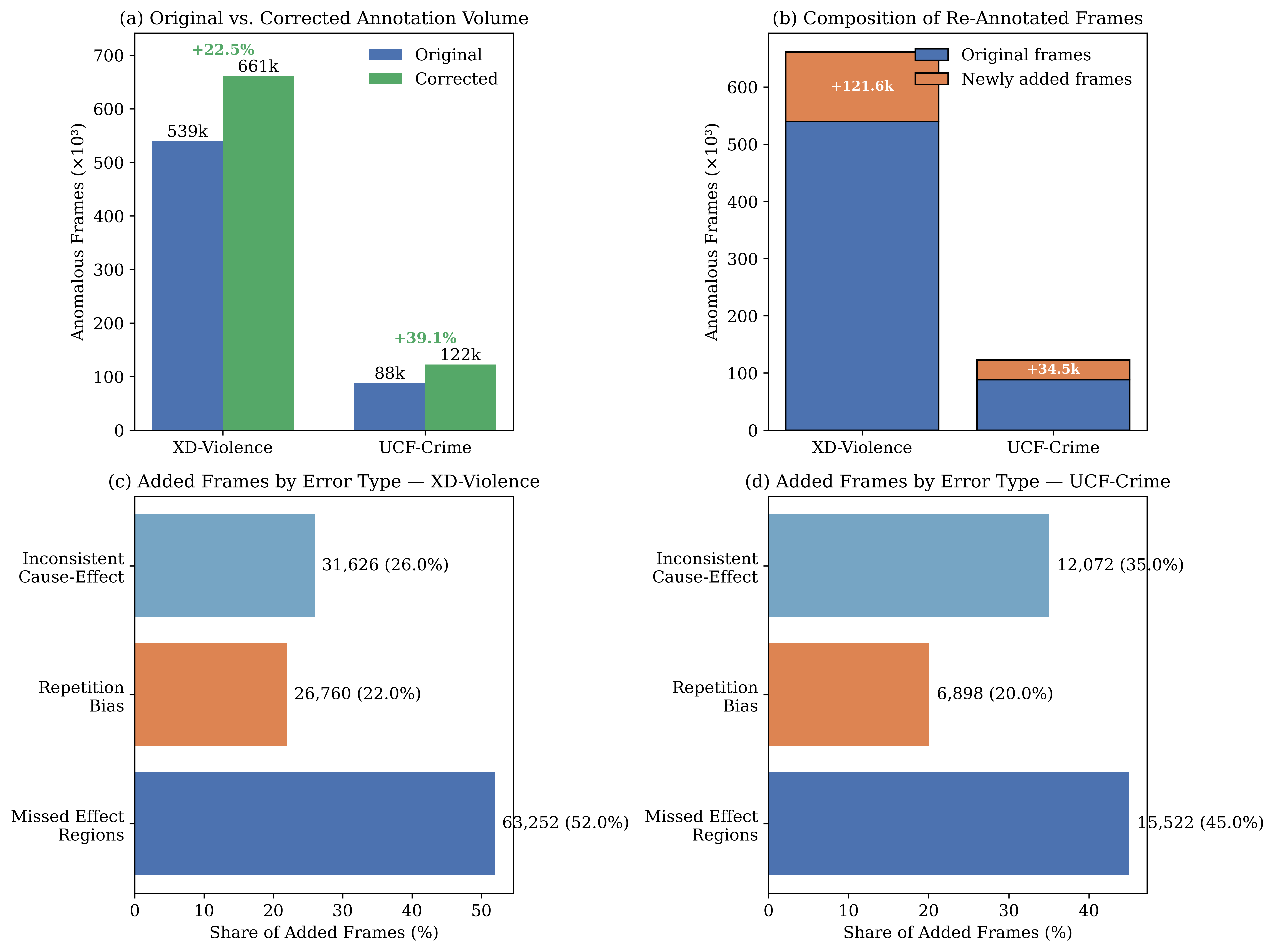}
    \caption{Quantitative analysis of re-annotation changes under the proposed unified labeling protocol. (a) Original vs. corrected anomalous frame volumes. (b) Composition of re-annotated frames showing original and newly added portions. (c–d) Distribution of added frames across three identified annotation error types for XD-Violence~\cite{wu2020looklistenlearningmultimodal}  and UCF-Crime~\cite{sultani2019realworldanomalydetectionsurveillance} , respectively.}
    \label{fig:ano_stas}
\end{figure*}

To address these inconsistencies and establish a reliable ground truth, we adopt a unified annotation principle:
\begin{quote}
\textit{All frames depicting the cause, effect, temporally connected escalation, or the aftermath of a real-world anomaly are labeled as anomalous.}
\end{quote}
\noindent Under this protocol, both XD-Violence~\cite{wu2020looklistenlearningmultimodal} and UCF-Crime~\cite{sultani2019realworldanomalydetectionsurveillance}  are re-annotated to ensure complete cause-and-effect coverage, consistent labeling of repeated anomalous events, inclusion of escalation and post-incident aftermath, removal of partial temporal truncations, and annotation of subtle contextual anomalies that are not purely motion-driven. This correction increased anomalous frames in XD-Violence~\cite{wu2020looklistenlearningmultimodal} from 539{,}562 to 661{,}201 (+121{,}639 frames, \textbf{+22.5\%}), and in UCF-Crime~\cite{sultani2019realworldanomalydetectionsurveillance} from 88{,}263 to 122{,}757 (+34{,}494 frames, \textbf{+39.1\%}).The statistical overview of the original vs ours unified reannotation has been shown in~\ref{fig:ano_stas}. All experiments in this paper are reported on both the original and unified annotations for fair comparison.

\subsection{Structural Analysis of Re-Annotations}
\label{subsec:struc_ana}
\begin{figure}[H]
\centering
\captionsetup{font=normalsize, skip=6pt}
\includegraphics[width=0.95\textwidth]{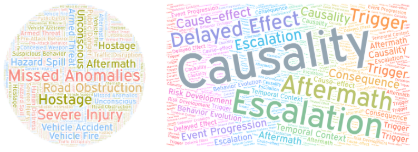} 

\caption{
Structural analysis of annotation distributions under the proposed unified protocol.
(a) Anomaly perception word cloud illustrating semantically enriched anomaly categories introduced through re-annotation, including subtle, contextual, and effect-based states that were under-represented in the original benchmarks.
(b) Temporal relation word cloud highlighting causal, escalation, and aftermath relations, emphasizing the necessity of modeling complete cause-effect chains rather than isolated trigger events.
Together, these visualizations provide linguistic and structural evidence that the original annotations are biased toward trigger-centric labeling, while our unified protocol captures temporally consistent and semantically grounded anomaly states.
}

\label{fig:anno_overview}
\end{figure}

To further characterize the structural differences between the original and unified labels, we present word-cloud visualizations of anomaly taxonomies and temporal-relation terms in Fig.~\ref{fig:anno_overview}. The anomaly perception word cloud highlights previously under-represented semantic states, while the temporal relation cloud emphasizes the necessity of modeling full cause-and-effect chains. Together, these analyses provide quantitative and linguistic evidence that the original annotations systematically under-represent semantically valid anomalous states.

\subsection{Annotation Procedure and Quality Control}
\label{app:annotation_procedure}

The unified re-annotations described in Sections~\ref{sec:ANOT_protocol}--\ref{subsec:struc_ana}%

were produced through a \emph{fully manual} process. No vision--language 
model, automatic relabeling system, or model-in-the-loop pseudo-labeling 
was used at any stage. This design choice is deliberate: because 
Cog-VADU is itself a VLM-based method, using any VLM to generate or 
assist the ground-truth labels would risk circular evaluation, where 
the labels are biased toward the same model family being evaluated. 
Manual annotation ensures that the corrected ground truth is 
\emph{independent} of the models under test.

\paragraph{Annotation protocol.}
Annotators followed the fixed unified principle stated in 
Section~\ref{subsec:principle}: all frames depicting the cause, effect, 
temporally connected escalation, or aftermath of a real-world anomaly 
are labeled as anomalous. To operationalise this principle 
consistently, each video was reviewed for three specific extensions to 
the original labels: (i)~repeated anomalous events that were annotated 
only at first occurrence, (ii)~effect-based anomaly continuations 
following a causal trigger, and (iii)~semantically anomalous contextual 
states that are not purely motion-driven. Decisions were made on the 
basis of \emph{temporal continuity and causal consistency} with the 
labeled event, rather than on visual saliency or model predictions.

\paragraph{Verification and bias-reduction safeguards.}
To reduce labeling errors and subjective bias, the annotations were 
subjected to the following safeguards:
\begin{itemize}
    \item \textbf{Multiple inspection passes.} Each re-annotated video 
    was reviewed in more than one pass, with the second pass 
    specifically checking temporal boundaries (start/end frames of each 
    extended segment) and the consistency of repeated-event labeling.
    \item \textbf{Protocol-driven decisions.} All corrections followed 
    the fixed causal-chain protocol rather than ad-hoc judgement, 
    ensuring that similar anomaly structures (e.g., cause~$\rightarrow$~effect~$\rightarrow$~aftermath) 
    are labeled consistently across videos and across the two datasets.
    \item \textbf{Model-independence.} Because no model predictions 
    were consulted during annotation, the corrected labels do not 
    inherit biases from any particular detector or LVLM, avoiding 
    circular reinforcement of model-specific behaviour.
    \item \textbf{Dual reporting.} To allow readers to assess the 
    effect of re-annotation transparently, all experiments are reported 
    on \emph{both} the original and unified annotations 
    Sec.~\ref{subsec:quantitative} (Table~\ref{tab:ucfcrime_auc}) for Xd-Violence and (Table~\ref {tab:xdviolence_auc}) for UCF crime datasets rather than replacing the original benchmark.
\end{itemize}

These safeguards are intended to ensure that the performance 
differences observed under the unified annotations reflect genuine 
improvements in anomaly coverage rather than artifacts of the 
re-labeling process.


\section{Chain-of-Anomaly Detection Thought Prompt (CoADTP)}
\label{sec:more_ab}
\begin{figure}[H]
\centering
\includegraphics[width=1.0\textwidth]{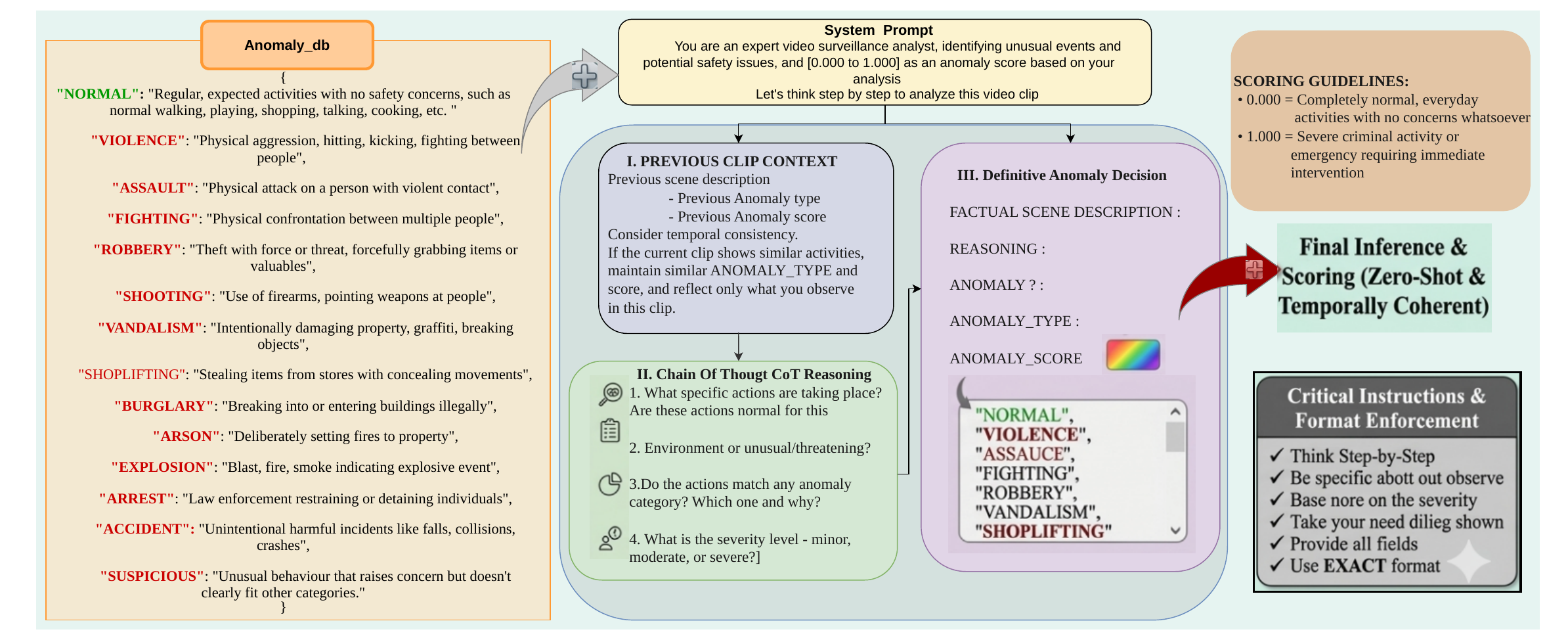} 
\caption{Workflow of CoADTP integrating anomaly database and temporal context and reasoning cues.}
\label{fig:my_cot_prompt_FIG}
\end{figure}

The CoADTP is illustrated in Fig.~\ref{fig:my_cot_prompt_FIG}. It is a step-wise prompting strategy that integrates three core components: (1) Chain of Thought style(Reasoning-based) (2) a textual anomaly knowledge base and (3) temporal feedback in the form of anomaly scores predicted and contextual information from previously processed clips for earlier segments. Incorporating these elements enables the model to perform temporally-aware reasoning, preserve contextual continuity across video segments, and refine anomaly predictions based on both accumulated evidence and domain-specific textual priors.

\noindent \textbf{Ablation on UCF-Crime}~\cite{sultani2019realworldanomalydetectionsurveillance} 
is reported in Table~\ref{table:Prompt_ab} to evaluate the contribution of each 
Cog-VADU component. 
We present the ablation along two complementary axes: (i)~\emph{single-component 
contributions} (rows 2--4) where each textual component is added in isolation above the 
base prompt, and (ii)~\emph{cumulative configurations} (rows 5--7) that progressively 
build up to the full pipeline. In single-component rows, numerical score smoothing is 
disabled so that each component's effect is measured in isolation.

Compared to a basic prompt (``Describe this scene and generate an anomaly score between 
0 and 1''), step-wise reasoning alone improves AUC from 0.7111 to 0.7548 (+4.37\%) and 
AP from 0.2150 to 0.2488 (+3.38\%), indicating more stable predictions. 
When evaluated in isolation, the Anomaly Database alone yields the largest 
standalone AP boost (AP $0.2150 \rightarrow 0.3392$, +12.42\%) with a modest AUC gain 
(+2.95\%), demonstrating that semantic priors are the strongest driver of anomaly 
\emph{localization}. Temporal Feedback alone yields a comparable AP boost 
(+11.53\%, AUC +3.28\%), confirming the value of prompt-level temporal continuity even 
without numerical score smoothing. Each of the three textual components therefore 
\emph{independently} improves over the base prompt; no component is redundant 
and they contribute along different metric axes (Reasoning $\rightarrow$ AUC; Anomaly DB 
and Temporal FB $\rightarrow$ AP).

In cumulative configurations, incorporating the anomaly database on top of reasoning 
boosts AUC to 0.7810 (+2.62\%) and AP to 0.3238. Adding temporal feedback further 
improves consistency across clips, yielding AUC 0.7888 (+0.78\%) and AP 0.3773 
(+5.35\%). Finally, the cross-modal re-ranking refinement stage achieves the 
best overall performance (AUC \textbf{0.8054}, AP \textbf{0.3808}), acting as a 
corrective post-processing mechanism that re-anchors textual rationales to visual 
evidence. Overall, the full Cog-VADU pipeline achieves 
a +9.43\% AUC and +16.58\% AP improvement over the base prompt baseline, 
with components contributing complementary rather than redundant gains.

\begin{table}[htbp]
\centering
\captionsetup{font=normalsize, skip=4pt}

\caption{Ablation study on UCF-Crime showing the contribution of each Cog-VADU
component. We report (i)~the base prompt, (ii)~each textual component 
evaluated in isolation above the base prompt (rows 2--4, with score smoothing disabled), 
and (iii)~cumulative configurations leading to the full pipeline (rows 5--7). 
Reasoning is the strongest AUC driver; Anomaly DB and Temporal FB are the strongest AP 
drivers; components are complementary rather than redundant.}

\label{table:Prompt_ab}

\renewcommand{\arraystretch}{1.0}
\newcommand{\rowstrut}{\rule[-0.8ex]{0pt}{2.8ex}}

\begin{tabularx}{\columnwidth}{Xcccccc}
\toprule

\textbf{Configuration} &
\textbf{Reasoning} &
\makecell{\textbf{Anomaly} \\ \textbf{Database}} &
\makecell{\textbf{Temporal} \\ \textbf{Feedback}} &
\makecell{\textbf{Refine-} \\ \textbf{ment}} &
\textbf{AUC} &
\textbf{AP} \\

\midrule

Base Prompt \rowstrut
& \xmark & \xmark & \xmark & \xmark
& 0.7111 & 0.2150 \\

\rowcolor{gray!12}
+ Reasoning only \rowstrut
& \cmark & \xmark & \xmark & \xmark
& 0.7548 & 0.2488 \\

\rowcolor{gray!18}
+ Anomaly DB only \rowstrut
& \xmark & \cmark & \xmark & \xmark
& 0.7406 & 0.3392 \\

\rowcolor{gray!12}
+ Temporal FB only \rowstrut
& \xmark & \xmark & \cmark & \xmark
& 0.7439 & 0.3303 \\

\rowcolor{gray!18}
+ Reasoning + DB \rowstrut
& \cmark & \cmark & \xmark & \xmark
& 0.7810 & 0.3238 \\

\rowcolor{gray!12}
+ R + DB + Temporal FB \rowstrut
& \cmark & \cmark & \cmark & \xmark
& 0.7888 & 0.3773 \\

\rowcolor{green!15}
\textbf{Full Cog-VADU} \rowstrut
& \cmark & \cmark & \cmark & \cmark
& \textbf{0.8054} & \textbf{0.3808} \\

\bottomrule
\end{tabularx}

\end{table}
\subsection{Choice of Large Vision Language Model}
\begin{figure}[!t]
\centering
\begin{minipage}[t]{0.56\textwidth}
\vspace{0pt}
\centering

\captionof{table}{AUC and AP (\%) comparison across LVLM model sizes and model families.}
\label{tab:model_sizes}

\renewcommand{\arraystretch}{1.5}
\begin{tabular}{lcc}
\toprule
\textbf{LVLM Model} & \textbf{AUC} & \textbf{AP} \\
\midrule
\rowcolor{gray!8}
VideoLLaMA3-2B & 75.09 & 24.39 \\
\rowcolor{gray!12}
InternVL2-8B & 76.26 & 31.01 \\
\rowcolor{green!15}
\textbf{VideoLLaMA3-7B} & \textbf{78.88} & \textbf{37.33} \\
\bottomrule
\end{tabular}

\vspace{1em}

\captionof{table}{Comparison of AUC and AP across different temporal window durations $T$ and number of sampled frames $N$.}
\label{tab:T_VS_N}

\scriptsize
\resizebox{0.75\linewidth}{!}{
\begin{tabular}{c|c|c|c}
\hline
\textbf{$T$ (sec)} & \textbf{$N$} & \textbf{AUC} & \textbf{AP} \\
\hline
\rowcolor{Gray!5}
5  & 5  & 78.15 & 34.02 \\

\rowcolor{Gray!10}
5  & 10 & 79.44 & 35.79 \\

\rowcolor{Gray!15}
10 & 5  & 77.65 & 29.03 \\

\rowcolor{Green!6}
10 & 10 & \textbf{80.54} & \textbf{38.08} \\
\hline
\end{tabular}
}
\end{minipage}
\hfill
\begin{minipage}[t]{0.42\textwidth}
\vspace{0pt}
\centering
\includegraphics[width=\textwidth]{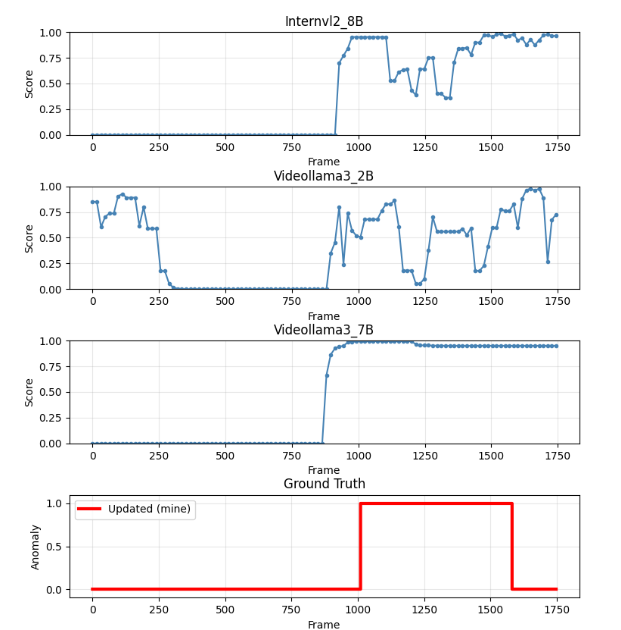}
\captionof{figure}{Output anomaly scores generated by InternVL2-8B \cite{chen2024internvl}, VideoLLaMA3-2B \cite{zhang2025videollama3frontiermultimodal}, and VideoLLaMA3-7B \cite{zhang2025videollama3frontiermultimodal}}
\label{fig:ty}
\end{minipage}

\end{figure}

To assess the generalizability and model agnosticism of CoADTP across different LVLM architectures, we evaluate multiple model sizes and families (Table~\ref{tab:model_sizes}). We first test InternVL2-8B, a stronger variant within the InternVL2 series. We then examine the VideoLLaMA3 family, including its 2B variants, all running at \texttt{bfloat16} precision~\cite{zhang2025videollama3frontiermultimodal}. Our findings indicate that VideoLLaMA3 models demonstrate substantially improved temporal reasoning and scene understanding compared to InternVL2-8B, resulting in more stable anomaly scores, as illustrated in Figure~\ref{fig:ty}. Notably, VideoLLaMA3-7B achieves the more promising performance, highlighting that improved cross-modal alignment and temporal comprehension contribute more than parameter count alone to anomaly detection effectiveness.
\subsection{Effect of Subsampling and Temporal Window}
Following ~\cite{ye2025veraexplainablevideoanomaly}, we adopt temporal-based uniform frame sampling for segment generation. 
Given a video $V$ with $F$ frames, we compute segment centers as
\[
C = \{ I_1,\; I_{d+1},\; I_{2d+1},\; \dots,\; I_{(h-1)d+1} \},
\]
where $d=16$ is the interval between centers and $h=\lfloor F/d \rfloor$ is the number of segments. 
For each center frame $I_{(u-1)d+1}$, we define a temporal window of $T$ seconds around it and uniformly sample N frames to form the segment $S_u$.

We further study the effect of varying the window duration $T$ and sampling rate as shown in ~\ref{tab:T_VS_N}. 
Empirically, using a 10-second window with a sampling rate of 1 FPS yields the best performance, which aligns with the training FPS of VideoLLaMA3
\definecolor{lightgreen}{RGB}{204,255,204}~\cite{zhang2025videollama3frontiermultimodal}
\newpage

\section{Faithfulness and Failure Analysis}
\label{app:faithfulness}
To substantiate the explainability claims made in the main paper, we 
conduct a focused qualitative study on three open-web videos that lie 
outside the training distribution of all evaluated baselines. These 
case studies serve three complementary purposes: 
\emph{(i)}~demonstrating that Cog-VADU's rationales are faithful to 
visual evidence; 
\emph{(ii)}~showing that the framework discriminates between visually 
similar but semantically different events; and 
\emph{(iii)}~documenting a failure mode of Cog-VADU and contrasting it 
with the failure modes of competing approaches.

\subsection{Symmetric Discriminative Cases}
\label{app:symmetric_cases}

We deliberately select two videos that form a \emph{symmetric pair}:

\begin{itemize}
    \item \textbf{Case 1 --- Visually anomalous, semantically normal:} 
    A professional arm-wrestling match (John Cena vs.\ Mark Henry) that 
    contains all surface cues commonly associated with anomalies 
    (physical struggle, audience reaction, post-match injury), yet is 
    contextually a sports event.
    \item \textbf{Case 2 --- Visually subtle, semantically anomalous:} 
    A nighttime recording of a missile interception over Tel Aviv, 
    which is visually ambiguous (bright lights against a city skyline 
    that could resemble fireworks) but corresponds to a genuine 
    safety-critical event.
\end{itemize}

Together, these cases establish that Cog-VADU's reasoning is genuinely 
\emph{context-driven} rather than biased toward a default visual 
pattern.

\begin{figure}[htbp]
\centering
\captionsetup{font=small, skip=6pt}
\includegraphics[width=\textwidth]{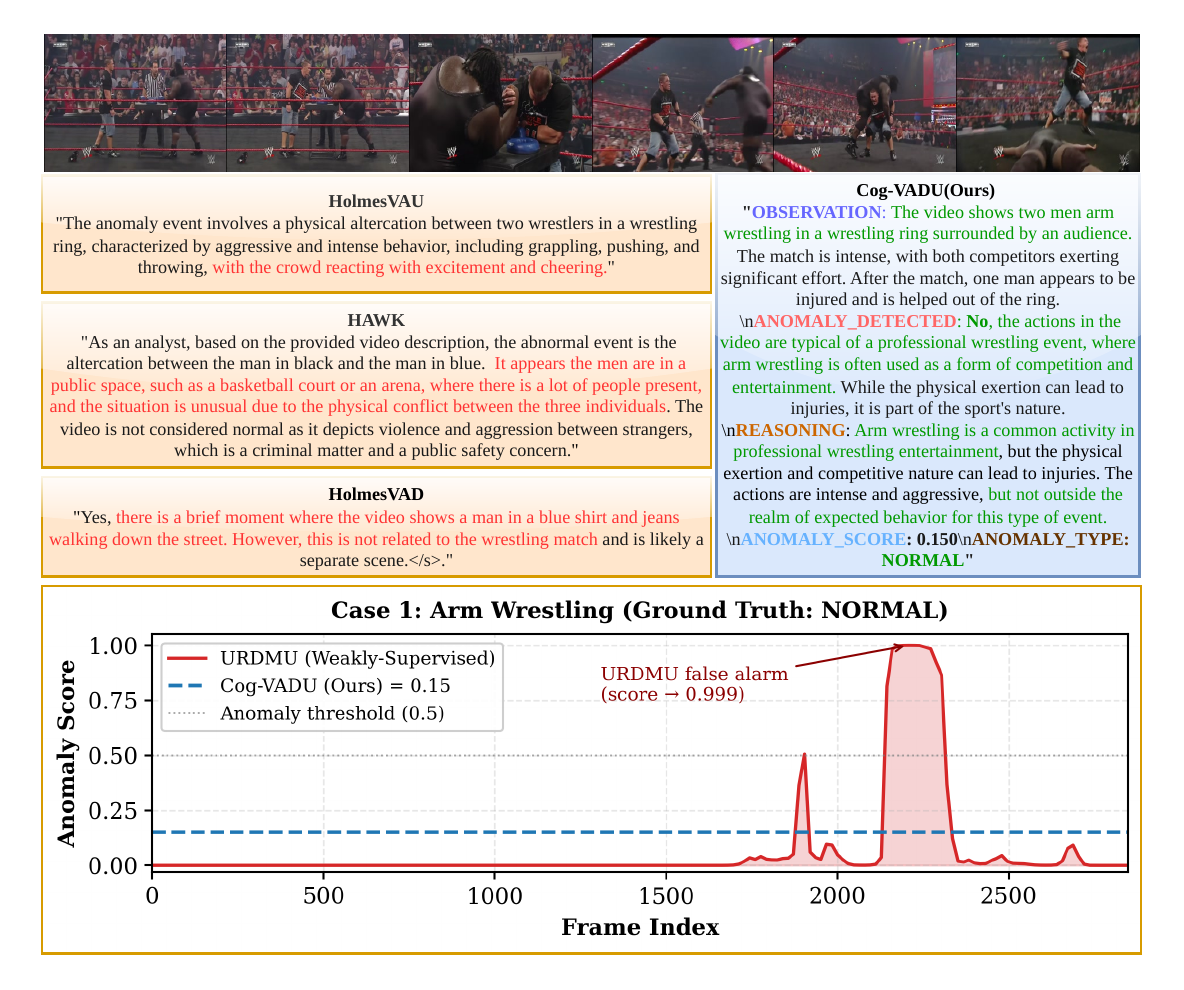}
\caption{\textbf{Case 1 --- Visually anomalous, semantically normal.} 
Comparison of four state-of-the-art VAD methods on an open-web arm 
wrestling video (ground truth: \textsc{Normal}). 
URDMU~\cite{zhou2023dualmemoryunitsuncertainty} produces a strong false alarm 
(peak $\approx 0.999$); 
HolmesVAD~\cite{zhang2024holmesvadunbiasedexplainablevideo} hallucinates an unrelated ``man in a blue 
shirt'' scene; 
HAWK~\cite{tang2024hawklearningunderstandopenworld} fabricates a basketball court context; and 
HolmesVAU~\cite{zhang2025holmesvaulongtermvideoanomaly} contradicts its own reasoning by labelling 
recognized wrestling activity as ``anomaly events.'' In contrast, 
Cog-VADU correctly identifies the professional-wrestling context and 
outputs a calibrated score of $0.15$ (\textsc{Normal}).}
\label{fig:case_wrestling}
\end{figure}

\begin{figure}[htbp]
\centering
\captionsetup{font=normalsize, skip=4pt}
\includegraphics[width=1.0\textwidth]
{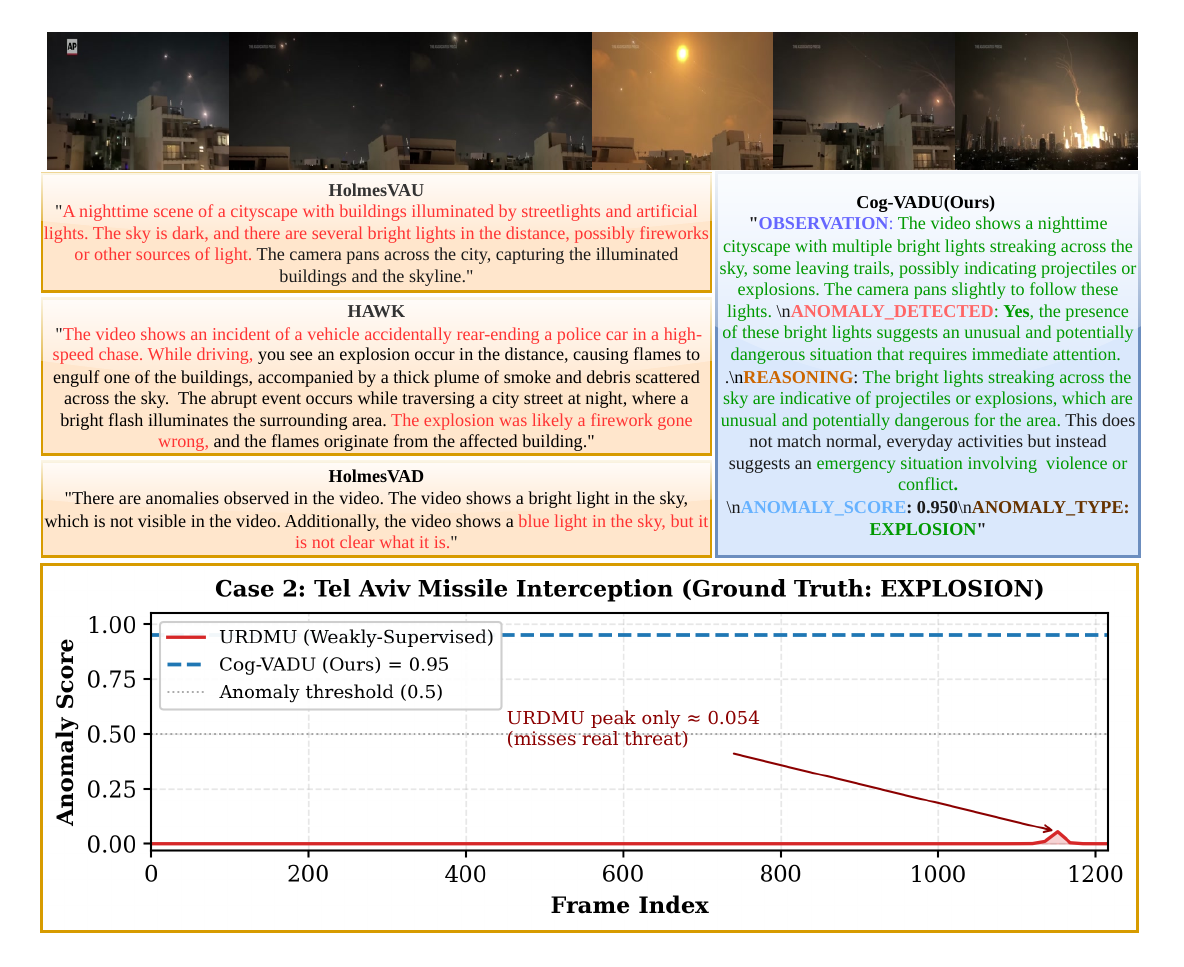}
\caption{\textbf{Case 2 --- Visually subtle, semantically anomalous.} 
Comparison on an open-web video of the Tel Aviv missile interception 
(ground truth: \textsc{Explosion}). 
URDMU~\cite{zhou2023dualmemoryunitsuncertainty} barely registers the event 
(peak $\approx 0.054$); 
HolmesVAD~\cite{zhang2024holmesvadunbiasedexplainablevideo} hallucinates a ``blue light in the sky''; 
HAWK~\cite{tang2024hawklearningunderstandopenworld} confabulates a vehicle accident and 
``firework gone wrong''; and 
HolmesVAU~\cite{zhang2025holmesvaulongtermvideoanomaly} misclassifies the projectiles as 
``fireworks or other sources of light.''}
\label{fig:case_telaviv}
\end{figure}

\subsection{Faithfulness of Cog-VADU Rationales}
\label{app:faithfulness_analysis}

In both cases, Cog-VADU produces grounded, evidence-aligned rationales 
that connect observation $\rightarrow$ reasoning $\rightarrow$ 
decision. For Case~1, Cog-VADU correctly identifies the 
wrestling-entertainment context (``arm wrestling is a common activity 
in professional wrestling entertainment'') and assigns a calibrated low 
score (\textbf{0.15}, \textsc{Normal}). For Case~2, Cog-VADU recognises 
the rapid trajectories as ``possibly projectiles or explosions'' and 
assigns a high score (\textbf{0.95}, \textsc{Explosion}). The 
rationales are factually consistent with the visual evidence and 
aligned with the final score.

\subsection{Failure Modes of Competing Methods}
\label{app:failure_modes_competing}

The two cases reveal three distinct failure modes in state-of-the-art 
baselines, demonstrating that current VAD methods fail in qualitatively 
different ways depending on their training paradigm:

\begin{enumerate}
    \item \textbf{Pattern-matching failure} 
    (URDMU~\cite{zhou2023dualmemoryunitsuncertainty}): The weakly-supervised method scores 
    the wrestling segment at $\approx 0.999$ purely from motion cues 
    (false alarm), while completely missing the Tel Aviv event 
    (peak $\approx 0.054$). This confirms that motion-statistical 
    approaches are blind to semantic context.
    
    \item \textbf{Rationale hallucination} 
    (HolmesVAD~\cite{zhang2024holmesvadunbiasedexplainablevideo}, HAWK~\cite{tang2024hawklearningunderstandopenworld}): 
    HolmesVAD fabricates ``a man in a blue shirt walking down the 
    street'' for the wrestling video, and HAWK confabulates ``a 
    vehicle rear-ending a police car in a high-speed chase'' for the 
    Tel Aviv footage - neither event exists. These are clear cases 
    where training-based MLLMs produce confident but ungrounded 
    explanations.
    
    \item \textbf{Reasoning--decision misalignment} 
    (HolmesVAU~\cite{zhang2025holmesvaulongtermvideoanomaly}): For Case~1, HolmesVAU acknowledges the 
    wrestling context yet still labels the events as ``anomaly 
    events,'' producing an internally contradictory output. For 
    Case~2, it misclassifies the missile interception as 
    ``fireworks,'' demonstrating that training-induced priors override 
    contextual evidence.
\end{enumerate}

\subsection{A Documented Failure of Cog-VADU}
\label{app:cog_vadu_failure}

\begin{figure}[H]
\centering
\captionsetup{font=small, skip=10pt}
\includegraphics[width=0.95\textwidth]{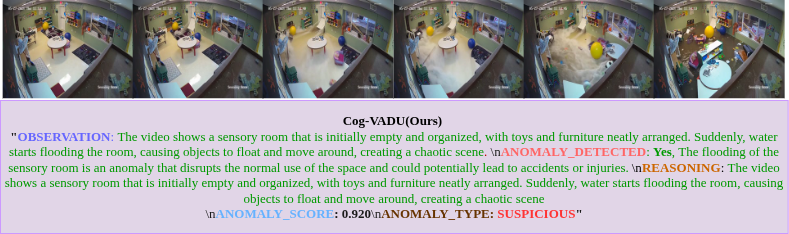}
\caption{\textbf{Failure case --- categorical labelling error under 
graceful degradation.} Cog-VADU is applied to an open-web video of a 
sensory room being flooded with water. The model produces a factually 
accurate \textit{description} (``water starts flooding the room, 
causing objects to float and move around''), correctly identifies the 
\textit{reasoning} (``not a normal activity and poses a safety 
concern''), and outputs a well-calibrated \textit{anomaly score} of 
$0.92$. However, the predicted \textit{anomaly type} is 
\textsc{Suspicious} instead of the more precise \textsc{Flood}. The 
root cause is structural rather than reasoning-related: our predefined 
anomaly database does not contain a \textsc{Flood} category, and the 
model gracefully defaults to the generic \textsc{Suspicious} class. 
This failure is correctable in a training-free manner by simply 
extending the textual anomaly database, in contrast to opaque 
training-induced failures observed in baselines.}
\label{fig:failure_flood}
\end{figure}

While Cog-VADU's rationales are largely faithful to the visual 
evidence, we have identified a recurring \emph{categorical labelling 
failure mode} that is informative for understanding the limits of the 
framework. Figure~\ref{fig:failure_flood} illustrates this on an 
open-web video showing a sensory room being flooded with water.

In this example, all reasoning sub-steps are correct: the description 
accurately identifies the flooding event (``water starts flooding the 
room, causing objects to float and move around''), the reasoning 
properly recognises the safety concern (``not a normal activity and 
poses a safety concern''), and the anomaly score is calibrated high 
(\textbf{0.92}). However, the final anomaly \emph{type} is 
misclassified as \textsc{Suspicious} instead of the more accurate 
\textsc{Flood} or \textsc{Accident}.

The root cause is structural rather than reasoning-related: our 
predefined anomaly database (Appendix~C, Fig.~10) does not contain a 
\textsc{Flood} category, and the model gracefully falls back to the 
generic \textsc{Suspicious} class; precisely the behaviour encoded 
in our prompt for ``unusual behaviour that raises concern but doesn't 
clearly fit other categories.'' This demonstrates two properties of 
Cog-VADU worth highlighting:

\begin{itemize}
    \item \textbf{Graceful degradation:} The framework correctly 
    detects, reasons about, and scores the anomaly even when its 
    vocabulary is incomplete.
    \item \textbf{Diagnosable failures:} Unlike hallucinatory failures 
    in baselines (Figures~\ref{fig:case_wrestling} and 
    \ref{fig:case_telaviv}), this failure is interpretable and 
    correctable simply by extending the anomaly database - without 
    any retraining.
\end{itemize}

\subsection{Summary}
\label{app:faithfulness_summary}

Taken together, the three case studies provide direct evidence that:
\emph{(i)}~Cog-VADU's rationales are grounded and consistent with 
predictions; \emph{(ii)}~the framework correctly disambiguates 
visually similar but semantically distinct events in both directions 
(suppressing false alarms on the wrestling case and recovering subtle 
anomalies on the Tel Aviv case); and \emph{(iii)}~competing 
training-based and pattern-based approaches fail in interpretable, 
diagnosable ways. The graceful-degradation behaviour observed in the 
flood case further indicates that Cog-VADU's failure modes are 
structural and addressable without retraining, a property not 
shared by competing supervised methods.

\section{Error Propagation Analysis}
\label{app:error_propagation}

A natural concern for any framework that propagates textual rationales 
across video segments is whether errors in early reasoning steps 
compound over time. This appendix presents a controlled 
rationale-corruption experiment that directly probes whether Cog-VADU 
exhibits error accumulation, along with an analysis of recovery 
dynamics and heavy-tailed cases.

\subsection{Experimental Setup}
\label{app:error_prop_setup}

We sample 20 UCF-Crime videos uniformly at random from the test set. 
For each video, we execute two runs with \emph{identical} visual 
inputs:

\begin{enumerate}
    \item \textbf{Baseline run:} Standard Cog-VADU with no 
    intervention.
    \item \textbf{Corrupted run:} After segment~0 produces its own 
    (correct) analysis, the temporal context $R_0$ passed to 
    segment~1 is \emph{replaced} with a deliberately incorrect 
    rationale. From segment~1 onward, the model runs normally but 
    with a contaminated reasoning chain.
\end{enumerate}

To cover realistic failure modes observed in competing methods 
(Appendix~\ref{app:failure_modes_competing}), we test three 
semantically distinct corruption types:

\begin{itemize}
    \item \textbf{Wrong category:} A normal-scene context is replaced 
    with a violent context (\textsc{Violence}, score $0.92$).
    \item \textbf{Inverted:} An anomalous-scene context is replaced 
    with a normal context (\textsc{Normal}, score $0.02$).
    \item \textbf{Hallucinated entity:} A fabricated rationale 
    describing a masked person with a weapon (\textsc{Robbery}, 
    score $0.88$).
\end{itemize}

For each segment $t$ we measure the absolute deviation
\begin{equation}
\Delta_t = |s_t^{\mathrm{corrupt}} - s_t^{\mathrm{baseline}}|,
\label{eq:deviation}
\end{equation}
and define \emph{recovery} as the first segment $\tau$ where $\Delta_t$ 
falls below $0.10$ and remains below that threshold for at least two 
consecutive segments. The choice of threshold $0.10$ reflects the 
typical scoring jitter we observe between independent runs of 
identical configurations.

\subsection{Recovery Statistics and Propagation Curves}
\label{app:error_prop_results}

\begin{figure}[!ht]
\centering
\captionsetup{font=small, skip=4pt}
\includegraphics[width=0.95\textwidth]{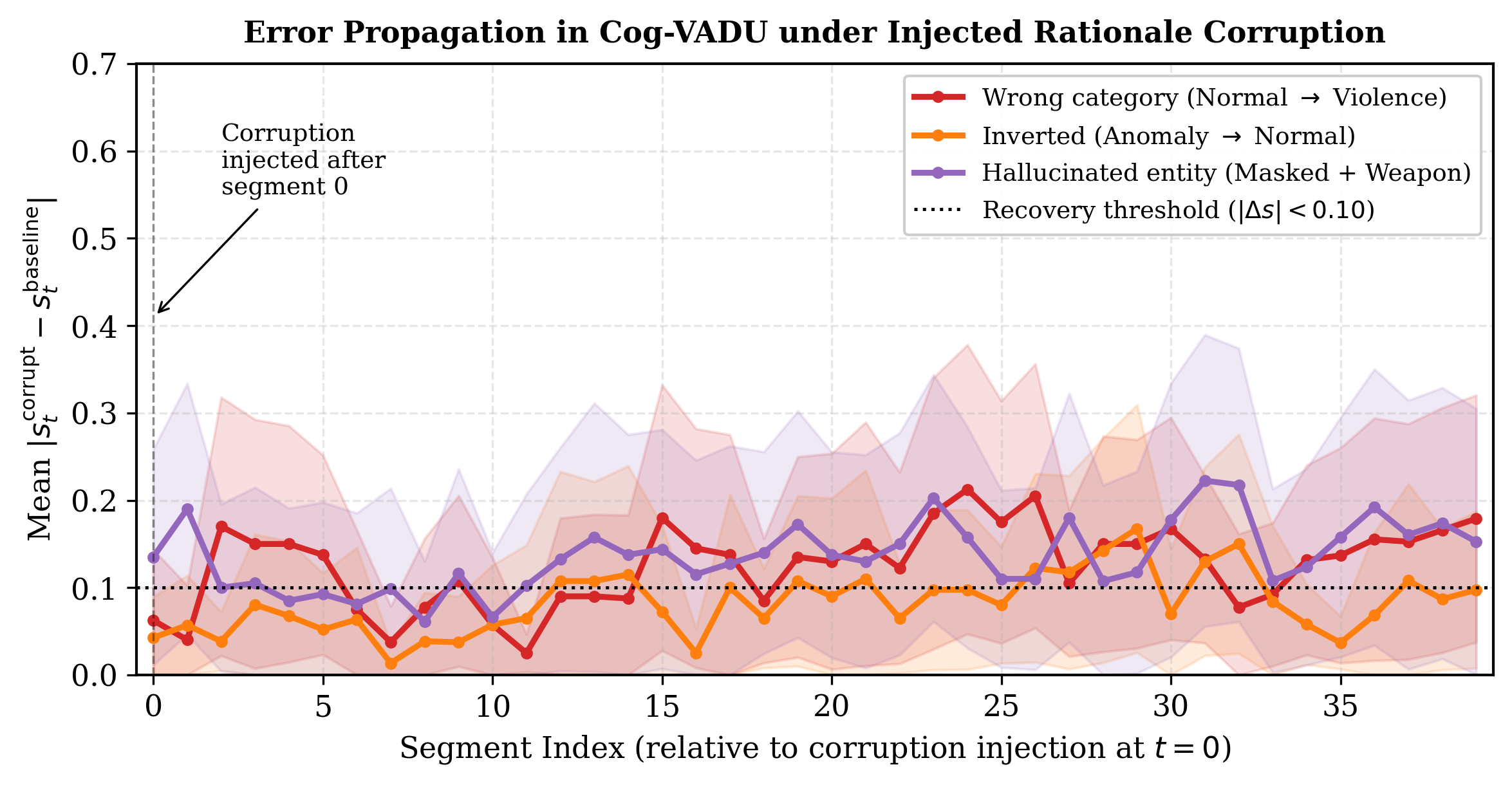}
\caption{\textbf{Error propagation under injected rationale corruption.} 
For each of three corruption types, we plot the mean absolute deviation 
between corrupted-run and baseline-run anomaly scores at each segment 
index after the injection point ($t=0$). Shaded bands denote 
$\pm 0.5\sigma$ across 20 UCF-Crime videos. Deviations remain bounded 
($|\Delta s| < 0.10$\,--\,$0.20$) rather than accumulating upward, 
demonstrating that Cog-VADU's per-segment visual grounding prevents 
long-range error propagation.}
\label{fig:error_prop}
\end{figure}

Table~\ref{tab:error_prop_stats} summarises recovery statistics across 
the three corruption types, and Figure~\ref{fig:error_prop} plots the 
mean propagation curves.

\begin{table}[!ht]
\centering
\captionsetup{font=small, skip=4pt}
\caption{Recovery statistics under three rationale corruption types on 
20 UCF-Crime videos. All $60/60$ runs recover within the test horizon; 
the median recovery time is 1 segment for every corruption type.}
\label{tab:error_prop_stats}
\small
\setlength{\tabcolsep}{4pt}
\renewcommand{\arraystretch}{1.2}
\begin{tabular}{l|c|c|c|c}
\toprule
\textbf{Corruption Type} & \textbf{Recovery Rate} & \textbf{Median $\tau$} 
& \textbf{Mean Max $|\Delta s|$} & \textbf{Mean Final $|\Delta s|$} \\
\midrule
Wrong category (Normal $\rightarrow$ Violence) & 20/20 (100\%) & 1 & 0.509 & 0.139 \\
Inverted (Anomaly $\rightarrow$ Normal)        & 20/20 (100\%) & 1 & 0.446 & 0.180 \\
Hallucinated entity (Masked + Weapon)          & 20/20 (100\%) & 1 & 0.512 & 0.142 \\
\bottomrule
\end{tabular}
\end{table}

\subsection{Key Findings}
\label{app:error_prop_findings}

Three observations follow from the results above:

\begin{enumerate}
    \item \textbf{Universal recovery.} Across all 60 
    video--corruption pairs (20 videos $\times$ 3 corruption types), 
    Cog-VADU recovers in $100\%$ of cases. The median recovery time is 
    \textbf{1 segment} for every corruption type, indicating that the 
    model self-corrects almost immediately after the injection point.
    \item \textbf{No upward drift.} The mean deviation curves remain 
    bounded throughout the entire 40-segment window 
    (Fig.~\ref{fig:error_prop}). Rather than compounding, deviations 
    oscillate around the recovery threshold 
    ($|\Delta s| \lesssim 0.10$\,--\,$0.20$), comparable to the 
    scoring jitter observed between independent runs of identical 
    configurations.
    \item \textbf{Visual re-anchoring.} The injected rationale is 
    overridden in subsequent segments because each segment is grounded 
    in fresh visual evidence and re-processed independently by the 
    LVLM. The temporal context biases the reasoning but does not 
    dictate it.
\end{enumerate}

\subsection{Heavy-Tailed Cases}
\label{app:error_prop_heavy_tail}

While the median recovery time is 1 segment across all corruption 
types, the \textsc{hallucinated entity} corruption exhibits a 
right-skewed distribution with mean recovery $5.4$ segments and 
standard deviation $10.6$. A small subset of videos therefore 
exhibits slower recovery than the median. On inspection, these cases 
correspond to videos in which the injected description 
(``masked person with a weapon'') is \emph{partially consistent} with 
the actual scene context, for example, low-light surveillance 
footage with occluded figures, making it harder for the model to 
contradict the injected context purely from visual evidence. This 
suggests that error recovery is fastest when the injected rationale 
sharply contradicts the visual content and slowest when ambiguous 
visual cues are compatible with the corrupted prior and a sensible 
and interpretable behaviour.

\subsection{Why Error Accumulation Is Bounded}
\label{app:error_prop_why}

We attribute the absence of error accumulation to two structural 
properties of Cog-VADU:

\begin{itemize}
    \item \textbf{Per-segment visual grounding.} At every segment, 
    the LVLM re-observes the raw frames and performs scene 
    understanding from scratch (Sec.~\ref{subsec:raw}). The prior 
    rationale enters the prompt as a textual hint, not as a 
    constraint, so it can be contradicted by current observations.
    \item \textbf{Cross-modal re-ranking in the refinement stage.} 
    Even when the textual rationale chain drifts briefly, the 
    refinement stage (Sec.~\ref{subsec:Analysis}) re-anchors 
    predictions to visual embeddings via FAISS-based retrieval, 
    providing an independent correction signal that further 
    suppresses any residual drift.
\end{itemize}

\subsection{Limitations}
\label{app:error_prop_limitations}

The analysis presented here demonstrates strong empirical robustness 
under a controlled 20-video pilot, but two practical limitations are 
worth noting. First, the experiment is conducted on UCF-Crime; 
very long-form surveillance footage ($>10$ minutes) may exhibit 
dynamics that we have not stress-tested here. Second, recovery is 
fastest when visual evidence sharply contradicts the corrupted 
rationale, so ambiguous scenes may exhibit longer-lived drift, as 
illustrated by the heavy-tailed cases in 
Sec.~\ref{app:error_prop_heavy_tail}.
\section{Computational Cost on Long Videos}
\label{app:computational_cost}

A natural concern is the computational cost of the multi-step 
reasoning framework, particularly on longer surveillance videos. 
This appendix reports inference cost for both modes in which Cog-VADU 
operates. All experiments were conducted on a single NVIDIA RTX 3090 
GPU (24\,GB VRAM, CUDA 12.6) using 
VideoLLaMA3-7B~\cite{zhang2025videollama3frontiermultimodal} in 
\texttt{bfloat16} precision.

Cog-VADU supports two evaluation modes that have fundamentally 
different cost profiles:

\subsection{Video-Level Reasoning Mode}
\label{app:cost_video_level}

This mode is used for anomaly understanding metrics 
(BLEU, METEOR, ROUGE-L, BERTScore) on HIVAU-70k. The model uniformly 
samples a fixed number of frames $N_f = 16$ from the \emph{entire} 
video regardless of duration and processes them in a single 
forward pass. Computational complexity is therefore approximately 
$\mathcal{O}(N_f)$ with $N_f$ fixed, making inference time 
approximately \emph{constant} in video duration.

In practice, inference time remains nearly constant even for long 
videos. For representative HIVAU-70k samples ranging from $43.8$\,s 
to $119.1$\,s, model inference required only $6$--$8$ seconds, while 
total processing time including frame extraction remained below 
$15$ seconds per video. This makes video-level reasoning practical 
for long-form surveillance footage: a 10-minute video and a 1-hour 
video incur essentially the same model-inference cost.

\subsection{Frame-Level Localisation Mode}
\label{app:cost_frame_level}

This mode is used for temporal anomaly detection metrics (AUC, AP) 
on UCF-Crime and XD-Violence. For fine-grained localisation, every 
16th frame is treated as a segment center, and a 10-frame window 
around it is analysed (Sec.~\ref{subsec:raw}). The complexity 
therefore scales linearly with video length, $\mathcal{O}(T)$, 
where $T$ denotes the number of frames in the video.

Inference is fully parallelizable across segments and 
across videos, so multi-GPU deployment scales linearly without 
coordination overhead.

\subsection{Discussion}
\label{app:cost_discussion}

The two modes reflect a practical trade-off between efficiency and 
temporal precision: \emph{sparse global sampling} (16 frames per 
video) enables efficient reasoning over long videos at the cost of 
no temporal localisation, while \emph{dense segment-based analysis} 
(every 16th frame as a center) provides accurate frame-level 
localisation at the cost of linear-in-duration runtime.

Two further properties bear on practical deployment. First, because 
Cog-VADU is \emph{training-free} and uses a frozen LVLM, no 
amortised training cost is incurred; the numbers reported above 
are the \emph{total} cost of deploying the system. Second, the 
segment stride $d$ in frame-level mode is a tunable parameter: 
deployments requiring near-real-time analysis can increase $d$ 
(sparser sampling) to trade temporal resolution for throughput. We 
leave a systematic study of stride-vs-accuracy trade-offs and 
backbone-size scaling (e.g., 7B vs.\ 2B variants) on long-form 
surveillance footage to future work.

\section{Additional Experimental Results}

We further evaluate our method on multiple widely used benchmarks, UCF Crime~\cite{sultani2019realworldanomalydetectionsurveillance}, XD Violence~\cite{wu2020looklistenlearningmultimodal}, and VANE ~\cite{bharadwaj2025vanebenchvideoanomalyevaluation}.

\subsection{Quantitative Results: Average Precision}

We additionally report Average Precision (AP), following the evaluation protocol used in prior work~\cite{wu2020looklistenlearningmultimodal}. 
Table~\ref{tab:ucfcrime_ap} compares Cog-VADU with state-of-the-art (SOTA) approaches on the UCF-Crime dataset~\cite{sultani2019realworldanomalydetectionsurveillance}. 
AP is particularly suitable for highly imbalanced datasets such as UCF-Crime, where anomalous events represent only a small fraction of the video timeline. 
As shown in the table, Cog-VADU achieves the highest AP among training-free approaches and surpasses several supervised baselines, demonstrating the effectiveness of reasoning-driven anomaly detection without task-specific training.

\begin{table}[htbp]
\centering
\captionsetup{font=normalsize, skip=5pt}
\caption{Average Precision (AP \%) on the UCF-Crime dataset~\cite{sultani2019realworldanomalydetectionsurveillance}. 
``Orig.'' denotes results evaluated on the original annotations, while ``Ours-Ann.'' reflects performance on our re-annotated split. 
Colored sections distinguish 
\textcolor{NavyBlue}{Weakly Supervised Non-Explainable}, 
\textcolor{ForestGreen}{Explainable}, 
and \textcolor{BurntOrange}{Training-Free Explainable} methods.}
\label{tab:ucfcrime_ap}

\renewcommand{\arraystretch}{0.9}

\begin{tabular}{lcc}
\toprule
\textbf{Method} & \textbf{AP (Orig.)} & \textbf{AP (Ours-Ann.)} \\
\midrule

\multicolumn{3}{l}{\textcolor{NavyBlue}{\textbf{Weakly Supervised Non-Explainable Methods}}} \\
\rowcolor{gray!12}UR-DMU~\cite{zhou2023dualmemoryunitsuncertainty} [AAAI-23] & 35.59 & 50.84 \\
\midrule
\multicolumn{3}{l}{\textcolor{ForestGreen}{\textbf{Explainable VAD Methods}}} \\
\rowcolor{gray!15}
VERA~\cite{ye2025veraexplainablevideoanomaly} [CVPR-25] & 32.83 & 45.12 \\
\rowcolor{gray!18}
Holmes-VAU~\cite{zhang2025holmesvaulongtermvideoanomaly} [CVPR-25] & \textbf{36.24} & \textbf{56.49} \\
\midrule
\multicolumn{3}{l}{\textcolor{BurntOrange}
{\textbf{Training-Free Explainable Methods}}} \\
\rowcolor{gray!21}
LAVAD~\cite{zanella2024harnessinglargelanguagemodels} [CVPR-24] & 27.27 & 38.38 \\
\rowcolor{gray!24}
\textbf{Ours} & \textbf{38.08} & \textbf{58.03} \\

\bottomrule
\end{tabular}

\end{table}

\subsection{Qualitative results: UCF and XD Violence}
\begin{figure}[htbp]
\centering
\captionsetup{font=normalsize, skip=4pt}
\includegraphics[width=1\textwidth]{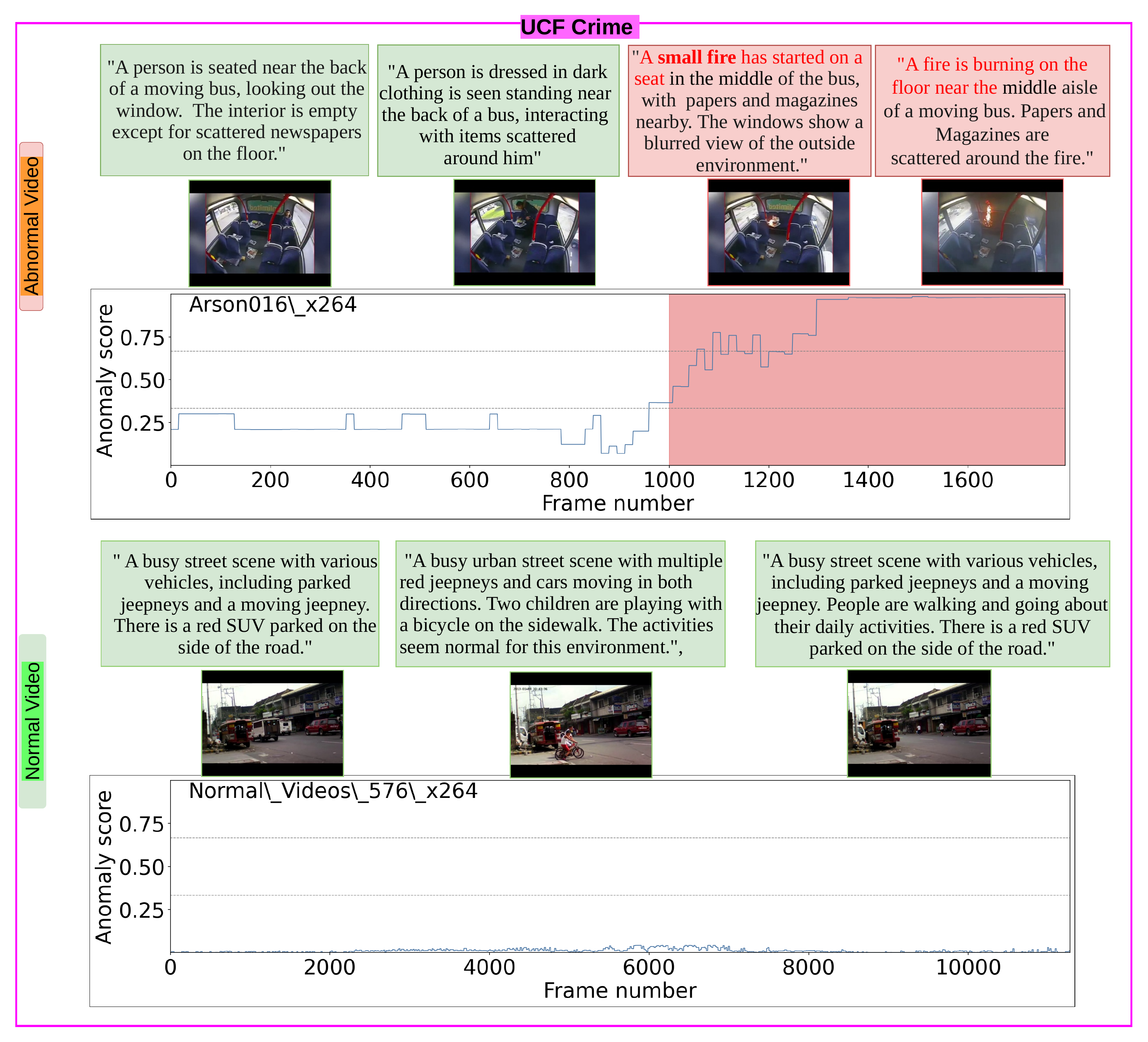}
\caption{Qualitative results of our Cog-VADU method on two test videos, from UCF-Crime~\cite{sultani2019realworldanomalydetectionsurveillance}. For each video, plot the frame-wise anomaly scores and show keyframes with their most relevant temporal summaries \textbf{(green for normal, red for abnormal)}. These examples illustrate alignment between predicted anomaly scores, visual content, and descriptions. Ground-truth anomalies are highlighted in red.}
\label{fig:more}
\end{figure}

\begin{figure}[htbp]
\centering
\captionsetup{font=normalsize, skip=4pt}
\includegraphics[width=1\textwidth]{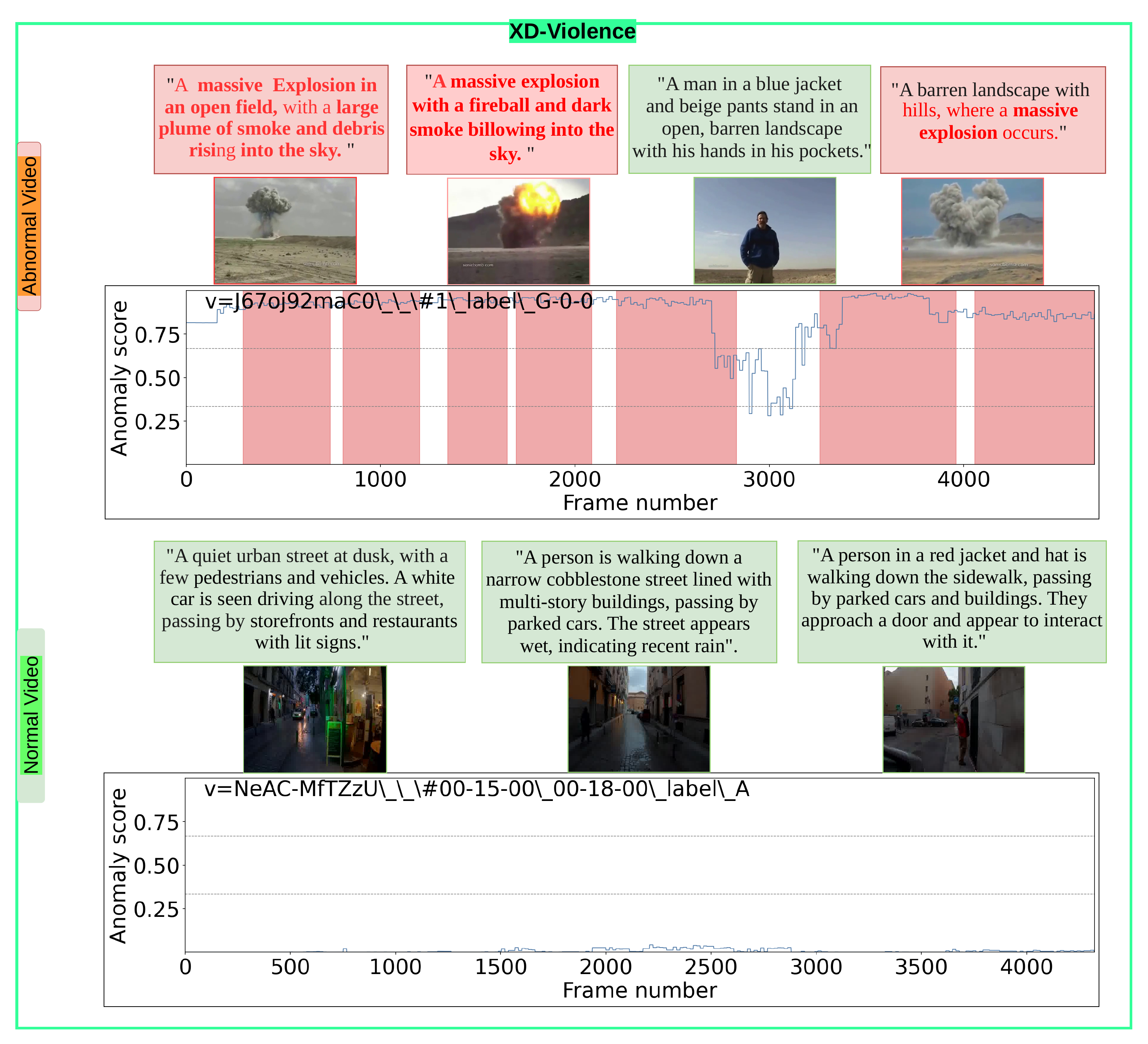}
\caption{Qualitative results of our Cog-VADU method on two test videos, from XD-Violence~\cite{wu2020looklistenlearningmultimodal}. For each video, plot the frame-wise anomaly scores and show keyframes with their most relevant temporal summaries \textbf{(green for normal, red for abnormal)}. These examples illustrate alignment between predicted anomaly scores, visual content, and descriptions. Ground-truth anomalies are highlighted in red.}
\label{fig:morexd}
\end{figure}

Figure \ref{fig:more} and \ref{fig:morexd} presents additional qualitative examples, with the top two rows taken from the UCF-Crime dataset and the two from the XD-Violence dataset~\cite{wu2020looklistenlearningmultimodal}. As shown, whenever an abnormal event occurs, i.e., an activity that deviates from the normal environmental context, the corresponding anomaly score produced by our Cog-VADU framework increases significantly. In contrast, for normal regions or normal videos (highlighted in green on the right), the anomaly scores remain low. Moreover, the temporal summaries generated by our method align well with the relevant video frames, further demonstrating the effectiveness of Cog-VADU in capturing and localizing anomalous events.

\subsection{Qualitative Results on the VANE Benchmark}

\begin{figure}[htbp]
\centering
\captionsetup{font=normalsize, skip=4pt}
\includegraphics[width=1\textwidth]{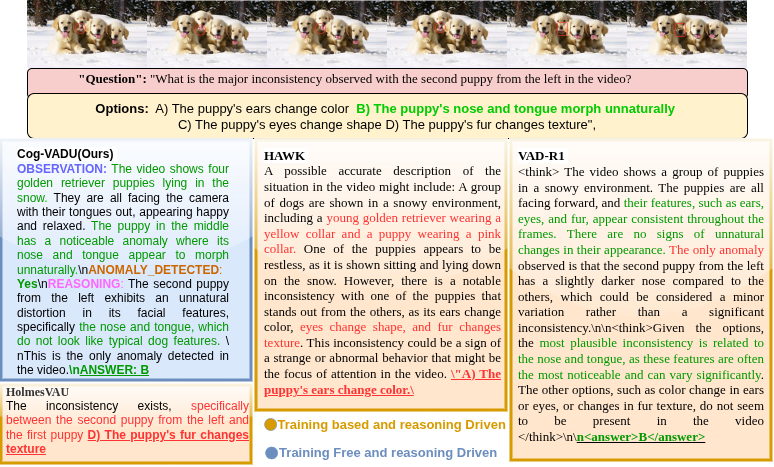}
\caption{
Qualitative comparison on a video reasoning benchmark. 
Our training-free framework \textbf{Cog-VADU} (blue) correctly identifies the anomaly by reasoning over temporal visual inconsistencies, detecting that the second puppy’s nose and tongue morph unnaturally across frames. 
In contrast, training-based VAD approaches (orange), including HAWK~\cite{tang2024hawklearningunderstandopenworld}, HolmesVAU~\cite{zhang2025holmesvaulongtermvideoanomaly}, and VAD-R1~\cite{huang2025vadr1videoanomalyreasoning}, either hallucinate unrelated changes (e.g., ear color or fur texture) or fail to localize the correct anomaly despite similar visual evidence. 
This example highlights the advantage of reasoning-driven anomaly detection in capturing subtle semantic inconsistencies that are not easily modeled by conventional training-based approaches.
}
\label{fig:morevane}
\end{figure}

Figure~\ref{fig:morevane} presents additional qualitative examples demonstrating the effectiveness of our method (Cog-VADU). In particular, we show results from the \textit{SORA} subset of the proposed VANE benchmark, which contains AI-generated synthetic videos exhibiting subtle and unconventional anomalies (e.g., unusual creature appearances or unrealistic environmental events) that go beyond traditional surveillance scenarios. These videos are generated using SORA, a state-of-the-art text-to-video generation model, and often appear highly realistic, making the underlying anomalies difficult to detect even for humans.
Automatically identifying such inconsistencies is therefore a challenging task and represents an important step toward detecting AI-generated or manipulated videos in real-world environments. Despite these challenges, our approach produces structured and grounded explanations that explicitly connect visual observations → cognitive anomaly reasoning → final decision concisely and coherently.

\noindent In contrast, HAWK~\cite{tang2024hawklearningunderstandopenworld} and Holmes-VAU~\cite{zhang2025holmesvaulongtermvideoanomaly} often fail to generate consistent reasoning chains, which can lead to incorrect predictions. 
VAD-R1~\cite{huang2025vadr1videoanomalyreasoning}, a reinforcement learning–based reasoning framework, produces the correct answer in this example but exhibits minor hallucinations in the reasoning steps.

\noindent Overall, these qualitative results highlight the advantage of Cog-VADU in producing accurate predictions together with concise and reliable reasoning.

\section{Limitations and Future Directions}

We introduce the task of \textit{Video Anomaly Reasoning}, enabling Multimodal Large Language Models (MLLMs) to perform deeper semantic analysis of anomalous events in videos in a zero-shot setting. While Cog-VADU demonstrates promising generalization from anomaly detection to anomaly understanding and reasoning, several limitations remain. In particular, the multi-step reasoning process increases inference time and introduces additional computational overhead. 

Future work will focus on improving efficiency through adaptive reasoning mechanisms and frame selection strategies that prioritize informative video segments. In addition, our findings highlight the importance of modeling temporal cause and effect relationships in anomaly detection. Motivated by this observation, we plan to develop a benchmark that explicitly captures such causal and temporal dependencies, providing a stronger evaluation platform for future research in video anomaly reasoning.
\end{document}